\documentclass[pmlr,twocolumn,10pt]{jmlr} 
\makeatletter
\ifdefined\LaTeXML 
\newif\ifmlhneedspmlr       \mlhneedspmlrtrue
\newif\ifmlhneedsstatements \mlhneedsstatementstrue
\newif\ifmlhfindings
\newif\ifmlhdemo
\newif\ifmlhanonymous       \mlhanonymoustrue
\def\mlhtrackname{Proceedings Track}

\def\mlhtrack#1{}

\def\floatconts#1#2#3{#2\label{#1}\begin{center}#3\end{center}}
\def\figureref#1{Figure~\ref{#1}}
\def\tableref#1{Table~\ref{#1}}

\def\acks#1{\paragraph*{Acknowledgments}#1}
\def\appendixref#1{Appendix~\ref{#1}} \fi
\mlhtrack{proceedings}

\newif\iffinal
\finalfalse  

\iffinal
    \ifmlhneedspmlr
      \jmlrvolume{XXX}
      \jmlryear{2026}
    \fi
    \ifmlhfindings \jmlrproceedings{}{ML4H 2026 - Findings Track}\fi
    \ifmlhdemo     \jmlrproceedings{}{ML4H 2026 - Demo Track}\fi
    \jmlrworkshop{Machine Learning for Health (ML4H) 2026}
\else
    \jmlrproceedings{}{Submitted to ML4H 2026: \mlhtrackname}
    \jmlrworkshop{Machine Learning for Health (ML4H) 2026}
\fi

\usepackage{booktabs}
\usepackage{siunitx}

\usepackage[switch]{lineno}

\usepackage{graphicx}
\usepackage{subcaption}

\theorembodyfont{\upshape}
\theoremheaderfont{\scshape}
\theorempostheader{:}
\theoremsep{\newline}

\title[Misaligned Cost Asymmetry in Open-Weight LLMs]{Misaligned Clinical Risk Classification and Cost Asymmetry in Open-Weight Large Language Models}

\author{%
 \Name{Star Liu}\nametag{\thanks{Corresponding author.}} \Email{sliu197@jhmi.edu}\\
 \Name{Xiyu Ding} \Email{xding20@jhu.edu}\\
 \Name{Robert B. Barrett} \Email{rbarre16@jh.edu}\\
 \addr Biomedical Informatics and Data Science, Johns Hopkins University School of Medicine, Baltimore, MD, USA
 \AND
 \Name{Alberto Santamaria-Pang} \Email{alberto.santamariapang@microsoft.com}\\
 \addr Microsoft, Redmond, WA, USA \\ 
 Biomedical Informatics and Data Science, Johns Hopkins University School of Medicine, Baltimore, MD, USA
 \AND
 \Name{Nic Dobbins} \Email{nic.dobbins@nih.gov}\\
 \addr National Institutes of Health, Bethesda, MD, USA \\
 Biomedical Informatics and Data Science, Johns Hopkins University School of Medicine, Baltimore, MD, USA
 \AND
 \Name{Harold P. Lehmann} \Email{lehmann@jhmi.edu}\\
 \addr Biomedical Informatics and Data Science, Johns Hopkins University School of Medicine, Baltimore, MD, USA
}

\begin{document}

\maketitle

\ifmlhdemo\else
\begin{abstract}

How large language models (LLMs) integrate patient risk with clinical cost tradeoffs remains poorly understood. We investigated how four open-weight LLMs (Qwen-2.5-7B/32B and Llama-3.1-8B/70B) internally represent cost tradeoffs, how these representations relate to clinical predictions, and whether decisions shift as predicted by the specified cost direction and magnitude. Using a public diabetes dataset, we varied 11 false-negative (FN) to false-positive (FP) cost ratios across three phrasings and examined representations and behavioral outputs. Patient risk was linearly recoverable on par with conventional classifiers (AUC $\approx$ 0.83), and cost direction was recoverable in every model. However, representational shifts in cost direction tracked output changes only in the two larger models, and responses to cost magnitude were predominantly direction-agnostic. Only 2 of 12 model–phrasings showed both opposing responses to increasing FN versus FP costs and cost-correct ordering. Representationally, a direction fitted on one cost side did not invert when transferred to the other, as expected under mirror-symmetric encoding. These findings suggest that LLMs encode risk and cost information but do not reliably integrate them into cost-correct decisions. Clinical evaluations should therefore include tradeoff tests, phrasing sensitivity, and default operating points alongside predictive performance. 

\end{abstract}
\fi

\mlhneedsstatementstrue
\paragraph*{Data and Code Availability.} This study used the Pima Indians Diabetes dataset, publicly available from the UCI Machine Learning Repository. Code repository will be released upon acceptance.

\paragraph*{Institutional Review Board (IRB).} This research does not require IRB approval. It uses only a publicly available, de-identified dataset.

\section{Introduction}
\label{sec:intro}

Medical diagnosis rests on evidence synthesis, probabilistic reasoning, and value-based decisions \citep{sackett_evidence_1996, ledley_reasoning_1959, keyes_evidence-based_2026}. The clinical decision made is not merely a probabilistic-estimation problem, but is also based on values\citep{pauker_threshold_1980, von_neumann_theory_1944}. Value assessments involve incorporation of contextual value judgements (e.g. ethical, moral, economic, and social constraints), along with probabilistic outcome estimates, toward a more holistic and optimal strategy. Thus, probability estimation integrated with value judgements lies at the heart of medical diagnosis. 

Where statistical and machine learning models presume symmetric costs \citep{elkan_foundations_2001}, clinical classification tasks reflect tradeoffs by the decision maker between false negative (FN) and false positive (FP) errors: a numerically low probability could represent “high” risk, if the human cost of FN is exceedingly high. In clinical decisions, FNs are usually perceived as worse than FPs, resulting in cost asymmetry \cite{newman-toker_burden_2024, newman-toker_diagnostic_2022, rosen_physicians_2000}, although perspective matters \citep{armstrong_patient_2019}. Similar probability of disease can warrant opposing actions, depending on the asymmetric costs of these two error types \citep{ledley_reasoning_1959, pauker_threshold_1980}. 

Computer-based decision support (CDS) is frequently used to aid in clinical decisions \citep{shah_making_2019}. A core consideration for adopting CDS is the alignment of the FN/FP tradeoff the CDS uses with institutional policy, even if these practices are implicit \citep{vickers_net_2016, wynants_three_2019}. Properly quantifying and incorporating tradeoffs into CDS development and evaluation is necessary for transparent and responsible deployment in the clinical context of use \citep{health_transparency_2024}.

Large language models (LLMs) are artificial intelligence systems, typically built on the Transformer architecture \citep{vaswani_attention_2023}, that are trained on massive corpora of data to emulate and generate human language \citep{brown_language_2020, zhao_survey_2026}. LLMs have increasingly been developed, evaluated, and used in CDS applications for medical diagnosis, prognosis, triaging, and reasoning tasks \citep{hager_evaluation_2024, xu_harnessing_2025, roeschl_assessing_2025, ullah_challenges_2024, schuemie_standardized_2025, ramaswamy_chatgpt_2026, goh_large_2024, lee_use_2026, bedi_testing_2025}. Moreover, LLMs have exhibited strong performance across information retrieval, synthesis, and predictions tasks when compared with human physicians, with increasingly comprehensive benchmarks to evaluate them \citep{goh_large_2024, bedi_testing_2025, jiang_health_2023, arora_healthbench_2025, van_veen_adapted_2024}. Recent evaluations raised concerns that existing benchmarks may no longer adequately distinguish model capabilities or reflect the demands of real-world clinical practice \citep{brodeur_performance_2026}.

To date, evaluation of LLMs in the clinical context has focused on their outputs. Such evaluation has reported failures with a clear cost-asymmetry. A recent study by \citet{ramaswamy_chatgpt_2026} showed that ChatGPT Health under-triaged clinical emergency vignettes (e.g. diabetic ketoacidosis, respiratory failure) in up to 52\% of patients, underscoring the need to incorporate asymmetric costs in LLM-based CDS. \citet{qazi_automation_2026} found that physicians exposed to erroneous LLM outputs frequently missed such errors, ending in reduced diagnostic accuracy. Such findings raise concern for whether clinician oversight can mitigate shortcomings in LLM-based CDS, and where automation bias may present greater risk. Distinguishing errors between those lacking cost information from those failing to incorporate cost is needed to advance LLM-based CDS solutions and evaluations. 

Rigorous evaluation of internal LLM mechanisms and failure modes is needed to minimize patient harm and maximize clinical benefit. Explicitly or implicitly, LLMs employ tradeoffs. Yet, there is limited literature on whether and how LLMs may internalize the value-theoretic principles toward optimal decisions. To the best of our knowledge, this study provides the first attempt to understanding how LLMs incorporate value-theoretic operations. Findings have implications for evaluating appropriate use of LLMs in downstream clinical decisions. We therefore ask: 1) does an LLM capture tradeoffs internally? 2) how does the internal capture of tradeoffs influence the final clinical prediction? 3) how does the decision boundary respond to the specified cost?

The rest of the paper is as follows. First, we discuss the decision analytic setup and relevant literature on clinical evaluations of LLMs and interpretability. The following section presents the study design and analytic methods. Subsequently, we present the results and findings addressing each of our primary questions. Finally, we discuss the implications for clinical LLM development, evaluation, and adoption.

\section{Background and related work} 
\label{sec:backg}

This work sits at the intersection of 3 relevant fields and directions of research: 1) decision analysis, 2) LLM application in the clinical domain, and 3) interpretability paradigms. In the subsections below, we discuss the background in each respective field and how they culminate to motivate this work.

\paragraph{Decision-theoretic foundations.} Our research question stems from the Pauker--Kassirer threshold-based decision-making framework \citep{pauker_threshold_1980, sox_measuring_nodate, phelps_focusing_1988}. The treatment threshold is $p^* = \mathit{Harm}/(\mathit{Harm} + \mathit{Benefit})$ [Equation 1]. 

A predictive model as a diagnostic test is often evaluated using receiver operating characteristic (ROC) curves. An ROC curve plots the true positive rate (TPR) against the false positive rate (FPR) across the entire range of predicted probability thresholds. The predicted probability threshold moves as the tradeoff moves. To maximize the test expected utility, both the subjective utilities and the probabilistic estimates must be used jointly{: $\text{slope at the ideal operating point} = (\mathit{Harm}/\mathit{Benefit}) \times (1/\text{Odds of outcome})$ [Equation 2]}.

\paragraph{LLM and clinical evaluation.} The evaluation of LLM-based CDS has used traditional metrics such as the area under the ROC curve (AUROC), gold standard chart review, qualitative human reviews, task-specific metrics and even another LLM \citep{bedi_testing_2025, jiang_health_2023, van_veen_adapted_2024, singhal_large_2023, bedi_holistic_2026}. However, these methods do not sufficiently aim at internal concerns, such as cost-sensitivity or how it impacts outputs. 

\paragraph{Interpretability.} “Interpretability" examines the “internal processes of AI models, moving beyond the evaluation of performance alone” \citep{bereska_mechanistic_2024}. There are 4 main approaches: 1) behavioral, 2) attributional, 3) concept-based, and 4) mechanistic (interventional) \citep{bereska_mechanistic_2024}. Behavioral interpretability relies on input and output; attributional interpretability relies on gradients; concept-based interpretability examines learned representations and patterns; and mechanistic interpretability tackles more granular analysis of neural-network neurons, connections, and precise computations to understand causal relationships governing model behaviors. Given our goal of understanding how tradeoffs are represented internally, we focused on  behavioral and concept-based observational interpretability. 

\paragraph{Desiderata.} To assess whether LLMs have the necessary components to perform decision-theoretic reasoning, let alone whether they do in fact perform such reasoning, we base our assessment on decision science and utility theory. Both the treatment threshold and expected-cost minimization together formalize how error-cost asymmetry should move a decision boundary. Any system that performs decision-theoretic reasoning must hold this cost structure internally, integrate it with probability to reach a decision, and move the threshold in the right direction by the right proportion. Such requirements are not immediately observable from outputs alone. We state three necessary conditions, such that a negative finding localizes the failure mode: \textbf{1) representation}, that direction and magnitude of a value tradeoff should be encoded and recoverable from activations \textbf{2) decision-layer use}, that specified cost tradeoff structure should influence the output decision, and \textbf{3) cost-sensitivity}, that the model should be cost-sensitive both directionally and in magnitude. Lastly, we require an additional constraint, that the model output under specified cost structure should be insensitive to paraphrasing. Violating this constraint reflects lexical influence rather than decision-theoretic reasoning. 

\section{Methods}
\label{sec:methods}

\paragraph{Dataset.} We used the Pima Indians Diabetes (PID) dataset, a publicly available medical dataset developed by the National Institute of Diabetes and Digestive and Kidney Diseases \citep{smith_using_1988,PID}. The dataset includes female patients aged 21 and over, with a binary diabetes status label. Predictors included age, pregnancy count, glucose concentration, blood pressure, skin thickness, insulin level, body mass index, and diabetes pedigree function score.

\paragraph{Dataset.} We used the Pima Indians Diabetes (PID) dataset, a publicly available medical dataset developed by the National Institute of Diabetes and Digestive and Kidney Diseases \citep{smith_using_1988,PID}. The dataset includes female patients aged 21 and over, with a binary diabetes status label, indicating diabetes onset within 5 years. Predictors included age, pregnancy count, glucose concentration, blood pressure, skin thickness, insulin level, body mass index, and diabetes pedigree function score.

\paragraph{Serialization and cost specification.} Each of the 768 patients from the PID structured dataset was serialized as a vector of 8 attribute/value pairs (Appendix A Table ~\ref{tab:apdA_serialization}). An example full prompt, including the system prompt, patient vignette, and the closing instruction is shown in Appendix A Table ~\ref{tab:apdA_full_prompt}. The LLM then classified vignettes in an experimental factorial design, with the following factors: model; Tradeoff Prompt Condition; Tradeoff Phrasing; and FN:FP ratio, totaling 180 classifications (\figureref{fig:study_design}). Factor details are presented in \tableref{tab:design-factors}. The \emph{cost\_only} prompt leverages standard language for describing FN and FP tradeoff, while the \emph{cost\_plus\_hint} prompt reflects the operational reality near the threshold for action, where cost tradeoff should break the tie. The lexical variants serve as a sensitivity analysis of the cost tradeoff expression.

\begin{figure}[t]
\floatconts
  {fig:study_design}
  {\caption{Study design. (a) 768 Pima vignettes with 4 models, 3 prompt conditions, 3 phrasings, and 11 FN:FP ratios. (b) One forward pass gives both the HIGH/LOW readout and the per-layer residual stream at the final prompt token. (c) Behavioral tests (ROC, mirror, coherence) and difference-of-means probes run on that shared pass. (d) Separating representation, use, and cost-correct use. Each model under 3 prompt conditions (\emph{cost\_only}, \emph{cost\_plus\_hint}, and \emph{tradeoff-free})  across 11 cost ratios gave 23 configurations (92 across 4 models). Two tradeoff paraphrases with 11 cost ratios across 4 models gave 88 configurations. In total, each patient vignette had 180 configurations.}}
  {\includegraphics[width=1\linewidth]{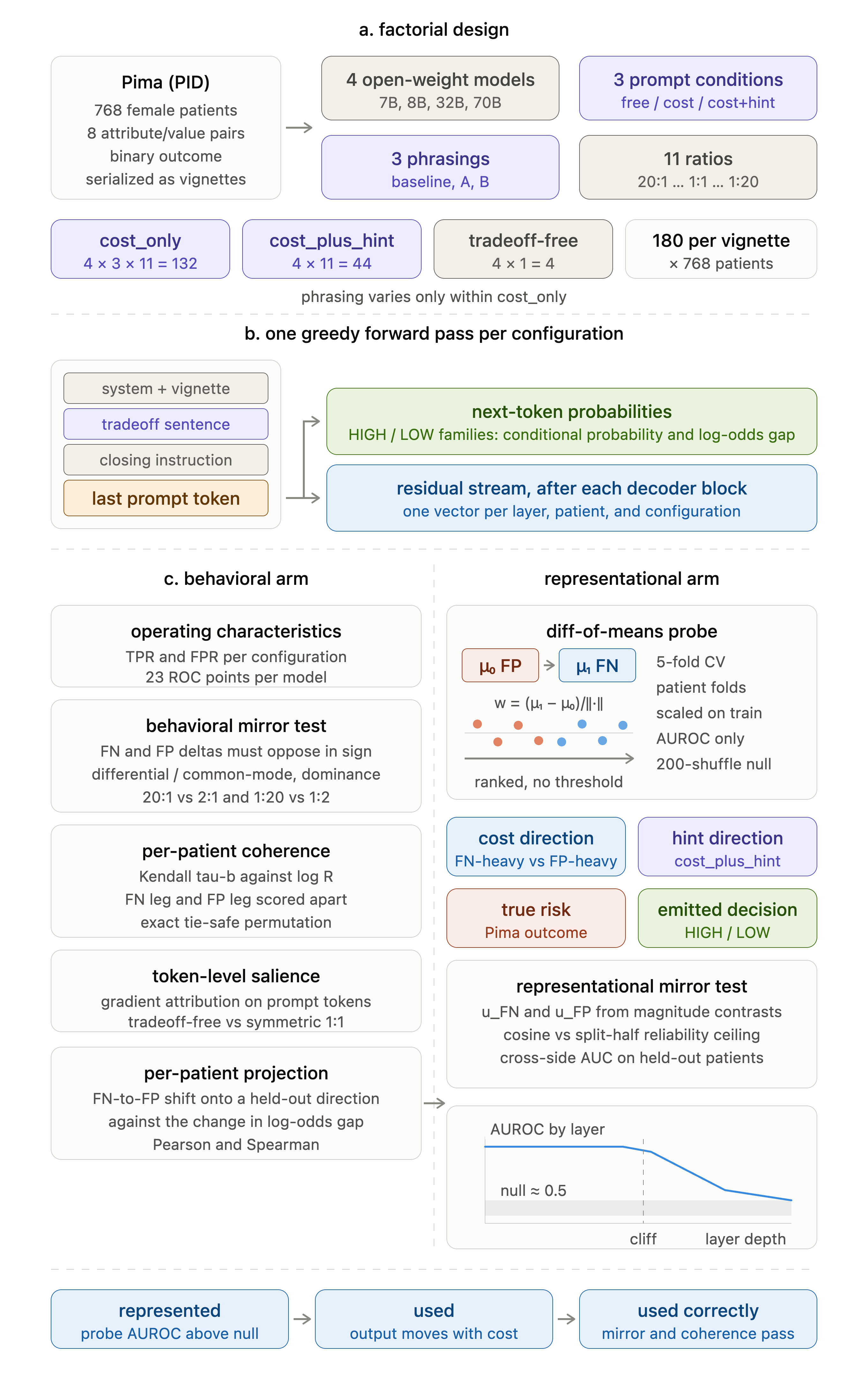}}
\end{figure}
\begin{table}[t]
\floatconts
  {tab:design-factors}
  {\caption{Design factors: models, tradeoff prompt conditions, tradeoff ratios, and decision-token families}}
  {\fontsize{9}{9}\selectfont
  \begin{tabular}{p{0.24\linewidth}p{0.65\linewidth}}
  \toprule
  \bfseries Factors & \bfseries Levels\\
  \midrule
  Open-weight LLM models & Qwen-2.5-7B, Llama-3.1-8B, Qwen-2.5-32B, Llama-3.1-70B\\
  \addlinespace
  Tradeoff prompt conditions &
    \textit{tradeoff-free}: No cost tradeoff\newline
    \textit{cost\_only}: Cost tradeoff only (``FN is $X$ times worse than a FP'').\newline
    \textit{cost\_plus\_hint}: Cost tradeoff + hint (``when uncertain, err towards high/low'')\\
  \addlinespace
  Tradeoff phrasing (paraphrases of the \textit{cost\_only} sentence) &
    \textit{Baseline}: ``The cost of a false negative (missing a patient who will develop diabetes) is [$X$] times the cost of a false positive (flagging a patient who will not develop diabetes).''\newline
    \textit{Variant A}: ``Missing a patient who will develop diabetes is [$X$] times as costly as flagging a patient who will not develop diabetes.''\newline
    \textit{Variant B}: ``A missed diagnosis is [$X$]-fold more serious than an unnecessary alert.''\\
  \addlinespace
  Tradeoff ratio combination &
    \textit{FN-heavy}: 20:1, 10:1, 5:1, 3:1, 2:1;
    \textit{Symmetric}: 1:1;
    \textit{FP-heavy}: 1:20, 1:10, 1:5, 1:3, 1:2\\
  \bottomrule
  \end{tabular}}
\end{table}

\paragraph{Behavioral extraction.} For each vignette and prompt configuration, we obtained the HIGH-versus-LOW prediction from the model's next-token probabilities, at the last prompt token position. Probabilities were aggregated across predefined families of single-token HIGH and LOW variants to account for differences in capitalization and tokenization. We reported $P(\text{HIGH} \mid \text{HIGH OR LOW})$ and log-odds gap $\log p(\text{HIGH}) - \log p(\text{LOW})$ where appropriate.

\paragraph{Operating-characteristic analysis.} For each Tradeoff Prompt Condition and Trade-off Ratio Combination, we calculated the true-positive rate (TPR) and false-positive rate (FPR). ROC curve was used to visualize how the specified relative costs shifted the model’s balance between sensitivity and specificity, resulting in 23 points in the ROC space per model.

\paragraph{Residual stream activation.} To examine the internal representations, we extracted the residual-stream activation after each decoder block’s attention and multilayer perceptron residual additions at the final prompt token during the same forward pass used to obtain the behavioral outcome. This extraction yielded one representation per layer, patient, and prompt configuration for the probing analyses described below. 


\paragraph{Token-level attribution.} We computed gradient $\times$ activation attributions \citep{shrikumar_learning_2017, li_visualizing_2016, simonyan_deep_2014}: at selected decoder layers, the inner product of each token's residual-stream activation and the HIGH versus LOW log-odds gradient. Appendix D Figure ~\ref{fig:apdD_salience} compares attribution patterns for the \emph{tradeoff-free} prompt and the symmetric 1:1 \emph{cost\_only} prompt. 

Additional details on behavioral outcome construction, operating-characteristic calculations, residual-stream activation extraction, and token-level attribution are provided in Appendix \appendixref{apd:extraction}.

\paragraph{Linear probes.} Complementing the behavioral analyses, we used linear probing, a concept-based interpretability method that asks what a model represents independently of what it outputs~\citep{cencerrado_no_2026, belrose_eliciting_2025}. We employed the difference-of-means probe~\citep{marks_geometry_2024, alain_understanding_2018}, following its use for recovering behavioral directions in LLM residual streams and motivated by the Linear Representation Hypothesis (LRH)~\citep{cencerrado_no_2026, park_linear_2024}. We collected layer-wise activation vectors for the two target classes (e.g., \emph{FN-heavy} vs.\ \emph{FP-heavy} prompts for the cost direction). We then computed the class centroids $\boldsymbol{\mu}_0, \boldsymbol{\mu}_1 \in \mathbb{R}^d$ on the training fold, took the unit-normalized difference direction $\mathbf{w} = (\boldsymbol{\mu}_1 - \boldsymbol{\mu}_0)/\lVert \boldsymbol{\mu}_1 - \boldsymbol{\mu}_0 \rVert$, projected each held-out activation onto $\mathbf{w}$, and classified it by comparing its projection to the midpoint of the two mean projections. This probe directly tested whether cost direction was encoded linearly that could be separated with no fitted parameters. \textbf{Prompt-side probe.} Within the \emph{cost\_only} and \emph{cost\_plus\_hint} conditions, we probed cost direction (\emph{FN-heavy} versus \emph{FP-heavy}, from the ten asymmetric ratios). \textbf{Ground-truth risk probe.} We probed the PID ground-truth label from the same per-layer activations used by the cost-direction probe. 
The same probe was run on the \emph{tradeoff-free} prompts to establish risk recovery independent of cost framing. Lastly, we probed the models' own emitted output.

Probes were evaluated with 5-fold cross-validation: a patient's activations, across every design configuration, were assigned to a single fold. Per-feature standardization was fitted on the training fold and applied to the test fold. We reported mean AUROC. Shaded bands showed the range across the 5 folds.


\paragraph{Per-patient projection analysis.} We examined whether patient-level representational changes under a cost-framing flip were associated with behavioral changes. At each layer, we projected each patient’s FN-heavy versus FP-heavy residual-stream shift onto a population direction estimated from the remaining patients. We then calculated Pearson and Spearman correlations between these projection scores and the corresponding changes in HIGH/LOW family log-odds, respectively. Calculation details and score interpretation are provided in Appendix \appendixref{apd:extraction}.

\paragraph{Cost-magnitude.} We evaluated whether the models used not only the direction of the specified cost tradeoff but also its magnitude. Let $z_p(r)$ denote the HIGH-versus-LOW family log-odds for patient $p$ under the FN:FP cost ratio $r$. Decision theory predicts that increasing the relative cost of a FN should lower the decision threshold for outputting HIGH and therefore increase $z_p(r)$. Conversely, increasing the relative cost of a FP should raise this threshold and decrease $z_p(r)$. We evaluated these predictions using two complementary tests: the mirror test and the coherence test. The mirror test assessed whether larger costs produced changes in the cost-correct directions, whereas the coherence test assessed whether responses were consistently ordered across the cost grid. Additional implementation and statistical-testing details are provided in Appendix \appendixref{apd:cost-magnitude}. \textbf{Mirror test} compared a larger and a smaller cost ratio within each cost direction. For the primary comparison, the FN-side contrast compared 20:1 with 2:1, whereas the FP-side contrast compared 1:20 with 1:2. Increasing the FN penalty should shift the model toward HIGH, whereas increasing the FP penalty should shift it toward LOW. Cost-correct magnitude use therefore requires a positive FN-side log-odds contrast and a negative FP-side contrast. We refer to this as a mirror test because the two sides should move in opposite directions. In contrast, a direction-agnostic response to the numerical magnitude alone would move both sides in the same direction. We repeated the analysis using 10:1 versus 2:1 paired with 1:10 versus 1:2, and 20:1 versus 5:1 paired with 1:20 versus 1:5. The mirror framework was applied to both behavioral outputs and residual-stream representations; we call the latter representational mirror test. \textbf{Coherence test} assessed whether the model's response was consistently ordered across the ratios within each cost direction. For each patient, we separately computed Kendall's $\tau_b$ between the HIGH-versus-LOW family log-odds and the ordered cost ratio across the five FN-side points (2:1 through 20:1) and the five FP-side points (1:20 through 1:2). The ratios were ordered so that a correct response should become more favorable to HIGH along both sides: the FN penalty increases along the FN side, while the FP penalty decreases along the FP side. Positive $\tau_b$ values on both sides therefore constitute a two-side positive signature.

All inference was greedy (temperature 0) with seed 0, using PyTorch [2.5.1+cu124] and transformers [5.5.4] on NVIDIA H100 GPUs; Llama-3.1-70B was sharded across two devices. Code will be made publicly available upon acceptance. 

\section{Results}
\label{sec:results}

\paragraph{Attribution.} As shown in Appendix D Figure ~\ref{fig:apdD_salience} from Lllama-3.1-70B-Instruct, the signals were scattered in the first layer. High impact tokens included “diabetes”, “pregnancies”, “plasma”, “glucose”, “148”, “Hg”, and “pedigree”. In the intermediate layer (layer 29), we saw shifts in both which tokens impacted the output and the degree. In the last layer, minimal attribution was observed. Compared to the \emph{tradeoff-free}, the 1:1 tradeoff prompt salience map exhibited similar attribution patterns. However, tradeoff specification shifted several tokens (e.g., “148” and “diabetes”) from HIGH (Appendix D Figure ~\ref{fig:apdD_salience}a) towards LOW (Appendix D Figure ~\ref{fig:apdD_salience}b). 

\paragraph{Ground truth probe.} Across all models, AUROCs hit a ceiling around 0.81-0.83 (Appendix D Table ~\ref{tab:apdD_risk_auc}). This performance ceiling (AUC = 0.83) aligned with standard supervised machine learning classifiers (logistic, boosting, and random forest) \citep{mienye_performance_2021}. Like the ground truth risk probe from the \emph{tradeoff-free} prompt, the ground truth risk probe from \emph{cost\_only} prompts with specified tradeoff had a similar range of AUROC performance. 

\paragraph{Behavioral (tradeoff direction).} For Llama-3.1-8B, the \emph{tradeoff-free} prompt was measured at 0.51 FPR and 0.88 TPR (\figureref{fig:frontier_grid}). Introducing a symmetric (1:1) tradeoff displaced every model toward LOW in family-sum log-odds, with median per-patient displacements of $-0.375$, $-0.250$, $-0.375$, and $-1.125$ for Qwen-2.5-7B, Llama-3.1-8B, Qwen-2.5-32B, and Llama-3.1-70B, respectively. In the two smaller models (Qwen-2.5-7B and Llama-3.1-8B), points did not form distinct clusters. In the two larger models, \emph{cost\_only} points formed distinct, though closely spaced, FN and FP clusters. Under \emph{cost\_plus\_hint}, points were more dispersed, but FN-heavy and FP-heavy points lacked clear intra-cluster separation. Family-sum log-odds distributions corroborate the ROC plot: log-odds cross zero under the symmetric 1:1 \emph{cost\_plus\_hint} prompt. Qwen-2.5-7B departed from the pattern shared by the other three models, with all points clustering in the high-FPR, high-TPR region of ROC space.

\begin{figure}[t]
\floatconts
  {fig:frontier_grid}
  {\caption{ROC curve with each point from a single tradeoff prompt over the entire dataset for all models. The blue star pins the performance from \emph{tradeoff-free} prompts. The blue circles are \emph{cost\_only} prompt results. The yellow squares are \emph{cost\_plus\_hint} results.}}
  {\includegraphics[width=1\linewidth]{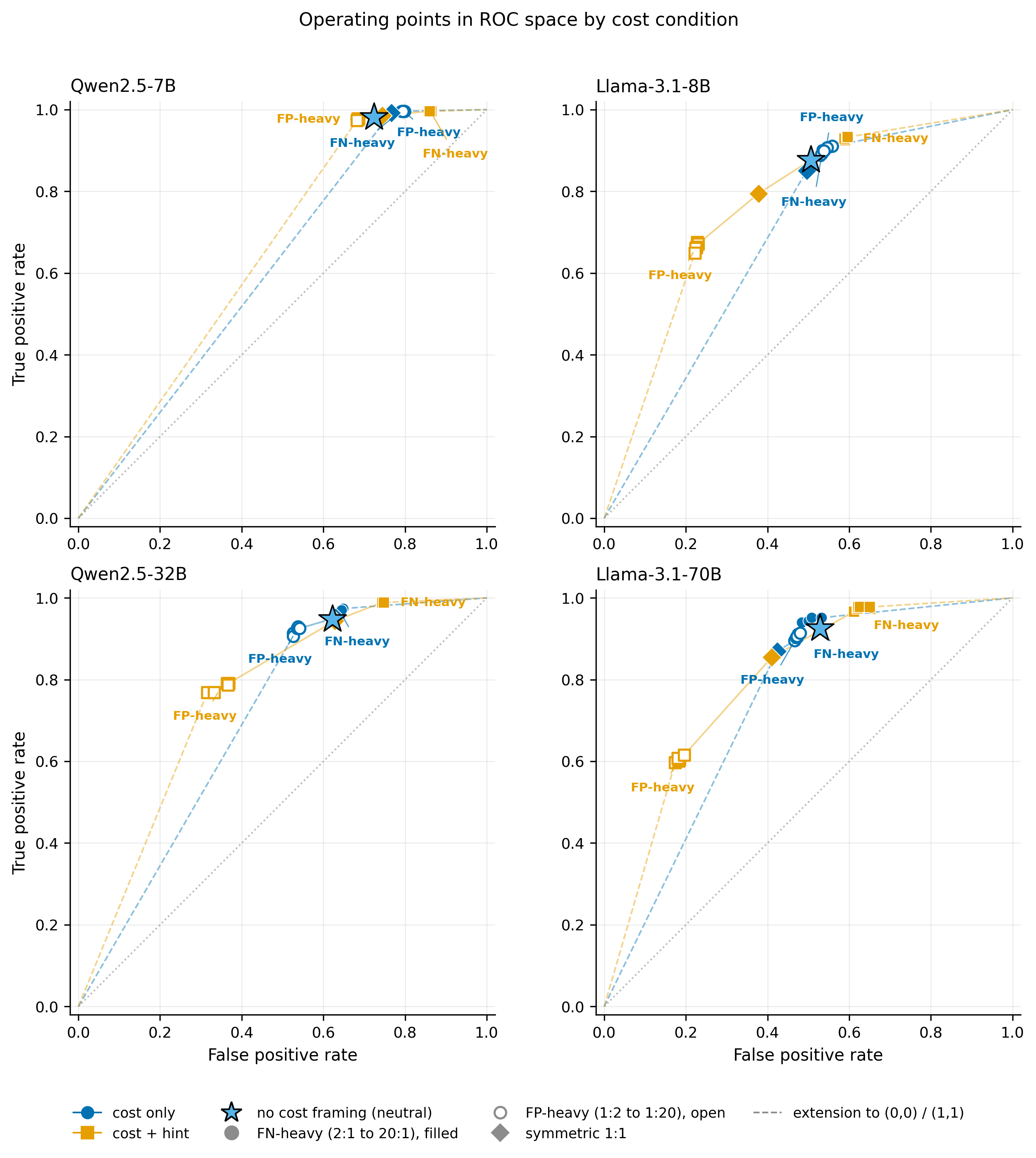}}
\end{figure}

\paragraph{Linear probes taken together.} Two features were expected from early layers largely encoding input text and detokenization. Thus, the cost-direction probe sits at AUC $=1.00$ due to lexical differences between the two cost tradeoffs. Late layers tend more to the output; thus, probing the model's own output approaches AUC $=1.00$ by design. For Llama-3.1-70B, the linear probe for cost-direction predicted which prompt (FN-heavy versus FP-heavy) was given perfectly in the first 30 layers (AUC = 1.00) (\figureref{fig:three_probe_panel}), where cost direction remained linearly decodable. The performance then dropped off a cliff by nearly 20\% throughout the intermediate to the last layer of the model, where the final output was tended to. Conversely, both the ground truth risk-probe and the model’s own predicted HIGH/LOW probe started low and then reached ceilings (AUCs of 0.83 and 1.00, respectively) right when the cost-direction probe drops off the cliff (layer 31). Since this decline coincides with the emergence of the decision, it is consistent with cost information being combined with patient information. Similar patterns were observed in the other models, each with its own cliff layer. Paraphrase variants of the original cost tradeoff sentence also recovered cost-direction via probes (with cliff regions). Only Llama-3.1-8B variants sustained high AUC throughout. Late-layer attenuation ordered consistently, with the original phrasing decayed the most, variant B intermediate, and variant A the least (Appendix D Figure ~\ref{fig:apdD_paraphrase_probes}). Probing \emph{cost\_plus\_hint} recovered the same layer-wise pattern at generally higher AUCs (Appendix D Figure ~\ref{fig:apdD_hint_probes}).

\begin{figure}[t]
\floatconts
  {fig:three_probe_panel}
  {\caption{Layer-wise cost-direction (blue), ground truth risk (orange), and model’s own HIGH/LOW probes' AUCs (green). Each line plots the mean AUROC with shaded bands reflecting the range across 5 folds. Vertical dotted line indicates the cliff region where the cost-direction probe performance begins to decline.}}
  {\includegraphics[width=1\linewidth]{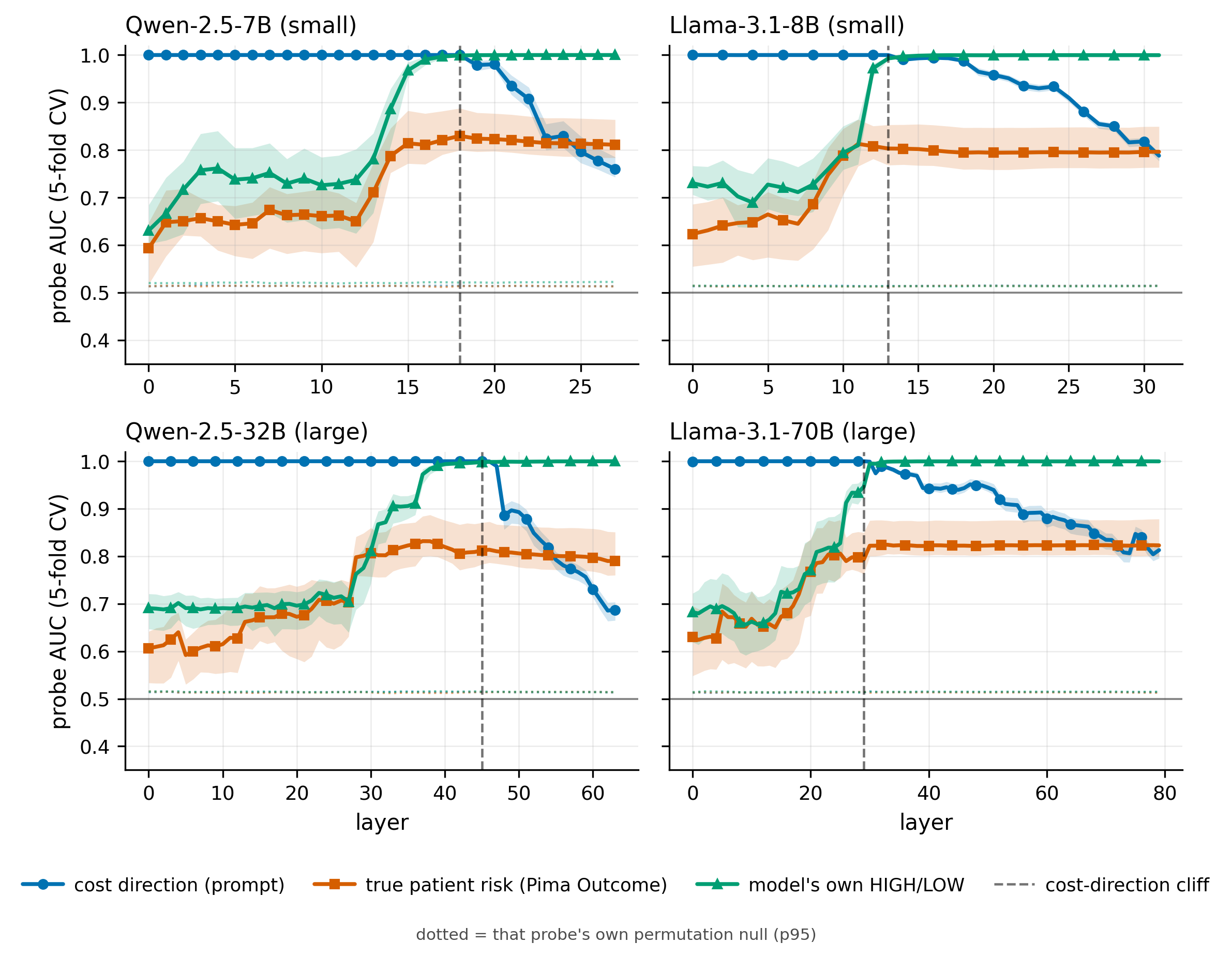}}
\end{figure}

\paragraph{Per-patient projection analysis} Scale-dependent patterns were observed for a 20-fold contrast between FN and FP (\figureref{fig:perpatient_projection}). Moderately negative correlation was found in Qwen-2.5-7B and Llama-3.1-8B; whereas, both Qwen-2.5-32B and Llama-3.1-70B showed higher correlation in the intermediate to late layers of the models. For Qwen-2.5-32B, the biggest jump occurred around layers 40-47 and remained high throughout the intermediate layers before a minor dip in the last 4 layers. For Llama-3.1-70B, there was an abrupt jump at layer 31 before declining and dropping in the final 3 layers. For Qwen-2.5-32B, the Pearson r (on logit) was as high as 0.73, and Spearman r (on logit) was as high as 0.72. For Llama-3.1-70B, the Pearson r (on logit) was as high as 0.56, and Spearman r (on logit) was as high as 0.66. All the remaining contrasts live in the Appendix D Figure ~\ref{fig:apdD_projection_behavior}. 

\begin{figure}[t]
\floatconts
  {fig:perpatient_projection}
  {\caption{Layer-wise per-patient activation projection versus behavioral delta per model, for a 20-fold contrast between FN and FP (20:1 versus 1:20). Both the Pearson (blue) and Spearman (orange) correlation curves for both log-odds and probability gaps are plotted separately. The vertical dotted lines mark the cliff region identified in the previous linear probe trajectory. Pearson and Spearman correlations between population-direction projection score, and, per-patient behavioral delta across all 768 patients. A high correlation indicates that patients whose representations shift more strongly along the cost direction also show larger logit-gap shifts under cost framing.}}
  {\includegraphics[width=\linewidth]{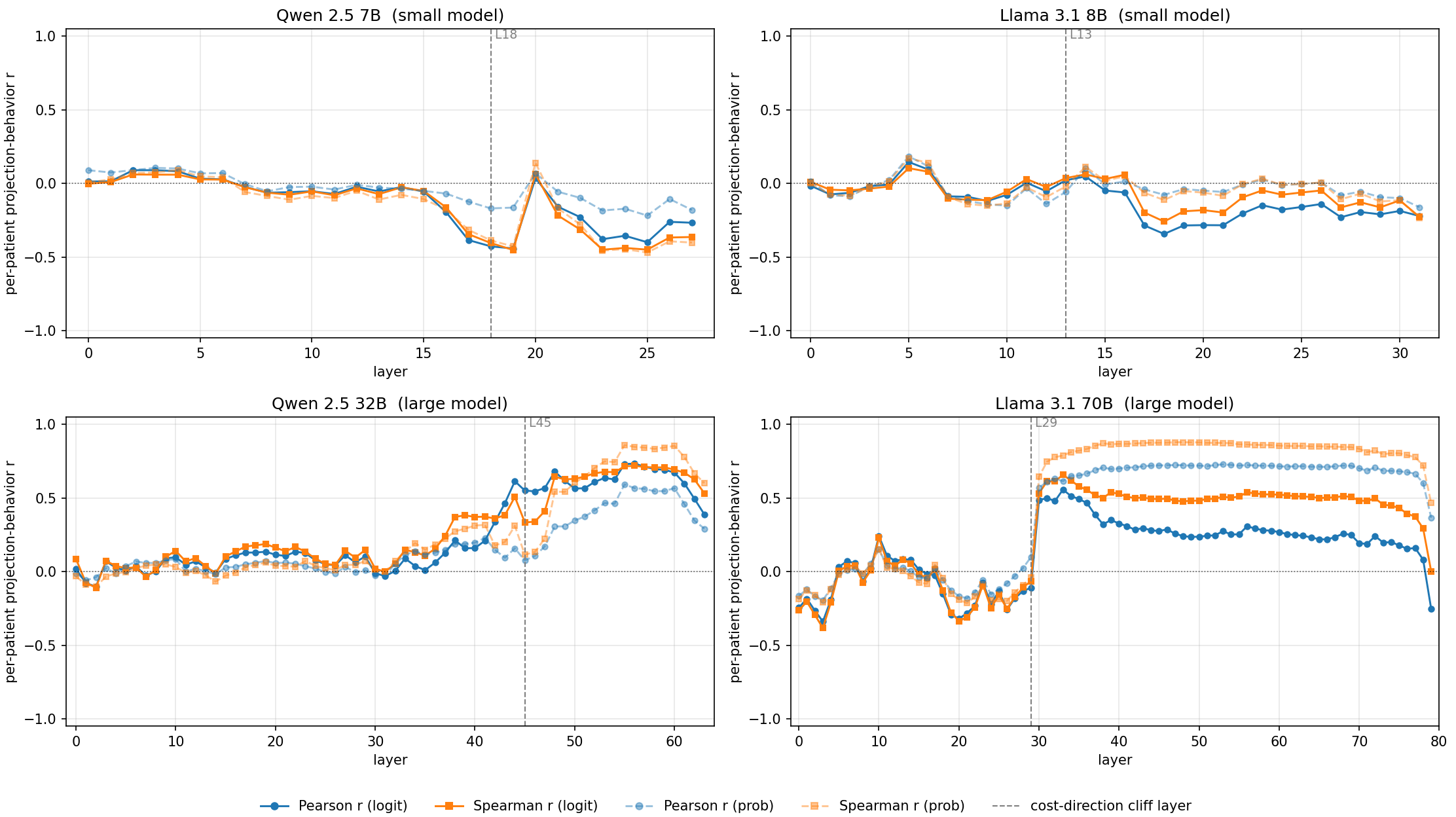}}
\end{figure}

\paragraph{Behavioral and representational mirror tests.} Among the 4 models with 3 types of phrasing (12 model-phrasing test), ten of which did not carry opposites signs for FN-side and FP-side deltas in the 20 versus 2 contrast (Appendix D Table ~\ref{tab:apdD_mirror_20v2}). Only 2 satisfied the mirror test: 70B-original (FN $\Delta = +0.713$, 99.9\% of patients moving in the right direction; FP $\Delta = -0.244$, 78.6\%) and 32B-original ($+0.459$, 94.4\%; $-0.672$, 85.8\%). Both Llama-3.1-70B and Qwen-2.5-7B variant Bs had high proportions of patients on the FP side correctly, but their FN side also had negative signs. Both smaller models fail under every phrasing, and both larger models fail under both paraphrases. Eight had small absolute differential and larger absolute common-mode (e.g. Llama-3.1-8B variant A and Qwen-2.5-7B), reflecting direction-agnostic response. Qwen-2.5-32B variant A had both FN and FP with the same sign. Qwen-2.5-32B variant B had a large differential, but the model moved toward LOW when FN cost increased. Other magnitude contrast mirror tests found that the 10 versus 2 reproduced similar results except for Llama-3.1-70B under paraphrases, where FN and FP deltas do carry opposite signs (Appendix D Tables ~\ref{tab:apdD_mirror_10v2} and ~\ref{tab:apdD_mirror_20v5}). Yet, Llama-3.1-70B only carried opposite signs under variant A for 20 versus 5 contrast.

\paragraph{Per-patient coherence test.} The mirror tests compared on an aggregate level and provides no insights into whether individual patients behave coherently. Median $\tau$ is positive on both sides (FN and FP) for both models (Llama-3.1-70B: FP $+0.80$, FN $+1.00$; Qwen-2.5-32B: FP $+0.74$, FN $+0.6$) with 77.7\% and 84.8\% of patients ordered cost-correctly on both sides (Appendix D Table ~\ref{tab:apdD_coherence}). 7B-variant-A has both sides positive $+0.11$, $+0.40$, but an order of magnitude weaker. All other 9 did not carry both positive $\tau$ with percent of patients with positive FN and FP $\tau$ ranging from 0.1\% to 37.8\%. Both Llama-3.1-70B and Qwen-2.5-7B variant Bs had digit-driven towards LOW. 32B-variant-B showed an inversion where the FN $\tau$ was negative with a 0 FP $\tau$.

\paragraph{Representational mirror test.} For Llama-3.1.-70B, FN and FP representations (under 20 versus 2 cost ratio contrast) were indistinguishable where the cosine score remained close to the common-mode ceiling in the first 30 layers and declined to around 0.8 in the remaining layers till the last (Appendix D Figure ~\ref{fig:apdD_repmirror_llama70b}). FN and FP specific components appear in the cliff region. However, the shared component accounted for 68 \ to 94\% of the representational response to cost magnitude. Lastly, a direction fit on one side (FN or FP) and applied directly to the other side ranks the held-out correctly with AUC between 0.65 and 1.00, but it never inverts. General patterns were observed in other models with model-specific variations (Appendix D Figures ~\ref{fig:apdD_repmirror_llama8b}-~\ref{fig:apdD_repmirror_qwen7b}). 

\section{Discussion}
\label{sec:discussion}

We aimed to understand how open-weight LLMs encode cost tradeoffs and whether that encoding is used when the model outputs a decision. This study spanned across 4 models and 3 semantically equivalent phrasings of the cost instruction prompt. Components of a cost-sensitive decision, patient risk and cost tradeoff were linearly recoverable from the residual stream at every scale. Decisions nonetheless track the stated tradeoff in only 2 of 12 model-phrasing tests in both the mirror and coherence tests. The representation exists across models of all scale but the downstream readout of the representation differs. 

\paragraph{An integration challenge.} Attribution graphs across all the patients showed early-layer attribution across most tokens, where they were reshaped into concentrated abstract intermediate states. This observed pattern was consistent with existing literature that highlighted early layer detokenization of input into local features before abstract construction in the intermediate layers \citep{sawant_mechanistic_2026, gurnee_finding_2023, lad_remarkable_2025}. Risk and cost direction co-decode across the entire pre-cliff plateaued including the cliff layer. Both were linearly recoverable at depths where the final output decision was still being computed. Early-layer decodability partly suggests lexical difference in the FN and FP prompts, and the key finding is joint decodability of cost and risk at the depth where the decision was being tended to. Cost direction and risk probe plateau well before the emitted decision becomes decodable, consistent with the tradeoff being carried as a feature separate from risk. Per-patient project analysis found correlation between representational changes and behavioral changes. Both the behavioral mirror and coherence tests suggest that the model access the tradeoff encoding but does not condition the model output on it. 

\paragraph{From a decision theoretic view.} Kassirer-Pauker threshold decision making requires that the threshold move in opposite directions under an increase in FN cost and under an increase in FP cost. Additionally, the magnitude of cost sensitivity should cohere with the size of the FN and FP tradeoff ratio. The mirror test findings suggest that magnitude response was direction-agnostic. The representational mirror test also agrees with the behavioral findings. A mirror-symmetric encoding should push the FN and FP directions oppositely, and the encoding should invert a prove fit on one side (FN or FP) when applied to the other. We did not uncover this structure despite differentiated representational states in the intermediate through late layers. Nonetheless, the FN and FP directions remain predominantly shared rather than opposed.

\paragraph{The default operating point.} A recommendation without a cost structure is underspecified. Thus, any model that emits a recommendation carries an implicit default operating point. Under a \emph{tradeoff-free} prompt, all four models lean towards predicting HIGH, with different points for each (Qwen-2.5-7B [TPR: 0.98, FPR: 0.72], Llama-3.1-8B [TPR: 0.88, FPR: 0.51], Qwen-2.5-32B [TPR: 0.95, FPR: 0.62], Llama-3.1-70B [TPR: 0.93, FPR: 0.53]). A stress test of ChatGPT Health found 52\% under triage of emergency vignettes \citep{ramaswamy_chatgpt_2026}. Moreover, triage shifted significantly when framing of the symptoms changed by family or friends. The acceptable tradeoffs likely vary over patient's demographics, comorbidities, and setting. The true acceptable tradeoff most likely deviates from the model’s implicit default threshold. This deviation is best framed as an expected utility penalization relative to the optimal operating point (Equation 2).

\paragraph{Implications for deployment.} The standard guardrail for clinical AI models has been human supervision, which assumes the human has relevant decision information readily accessible and that it may be efficiently retrieved. Traditional LLM-based CDS often circumvents these concerns by providing grounded textual evidence. This study presents a distinctly challenging form of explainability, where decision thresholds are not directly tangible artifacts, but instead internally derived. Automation bias has been reported under exactly these conditions \citep{qazi_automation_2026}. Further, relying on clinicians as overseers of models potentially exacerbates an existing alert-fatigue dilemma \citep{ray_alert_2026}. Benchmarks scored against a single implicit cost structure cannot detect when a model does not integrate cost ratio. Thus, evaluations for clinical deployments need tradeoff-conditioning tests and reporting of default operating point beyond ranking metrics alone. 


\paragraph{Implications for interpretability.} Linear decodability does not equate to use \citep{ravichander_probing_2021,elazar_amnesic_2021, belinkov_probing_2022, turpin_language_2023}, and LLMs can encode information that the outputs don't reflect \citep{orgad_llms_2024}. This work extends the represented-but-not-used finding to clinical cost reasoning with the caveat that the target is measurable against a theoretical value rather than a label. In the context of LRH \citep{park_linear_2024}, our cost-direction probe reached a ceiling near where the behavior is absent and remained linearly decodable where the models shifted to attend to the decision output. Yet the decision did not reflect cost-correct use in most conditions. Linear recoverability helped us understand information availability but not its precise functional role. The eventual objective is to identify causal interventions beyond observational analysis, to patch, and to reconstruct the circuitry binding against a represented tradeoff to the emitted decision output. 

\paragraph{Limitations.} First, the study findings were not causal. Activation patching is the intervention test where linear probing is limited. Nonetheless, this work provides the necessary first step to understand how tradeoff exists in representation before knowing where to assess causality. Second, the PID dataset is a limited testbed. Future work will include the Stanford INSPECT dataset across conditions and modality. Third, linear probes read the last prompt token position; thus, results do not rule out representation elsewhere. Fourth, tradeoff magnitude proportionality remains untested. The mirror test examined the sign, and the coherence test assessed ordering. Neither test measures whether the model tracks tradeoff size. Fifth, the mirror tests measured displacement and is informative only for patients with room to move in both the FN and FP directions. Sixth, future work will investigate reasoning and chain-of-thought models. More output tokens could enable probing analysis across positions and how cost information is carried into reasoning. Lastly, future work will establish benchmark datasets and an evaluation framework that includes explicit action decisions and cost framing for comprehensive sensitivity analysis.

\section{Conclusion}
\label{sec:conclusion}

Clinical tradeoffs are at the heart of clinical decision-making, and a model that emits an output decision carries an implicit tradeoff or operating point. We showed that 4 open-weight LLMs represent both patient risk and clinical cost tradeoffs in a linearly recoverable fashion. Yet, the deficit was one of value integration. Without understanding the internal mechanisms, a model could reach the right immediate answer by the wrong means, which could lead to adverse outcomes in the long run. Representation is not use, and evaluations should not treat the first as evidence of the second.





\acks{We thank Yuanji Han for helpful feedback on this work. \\ \\
Star Liu and Robert Barrett were supported by the National Library of Medicine
(NLM 5T15LM013979).}

\bibliography{ref}

\appendix




\setcounter{table}{0}
\renewcommand{\tablename}{Appendix~A~Table}
\renewcommand{\thetable}{\arabic{table}}
\renewcommand{\theHtable}{apdA.\arabic{table}}

\makeatletter
\@ifundefined{apdblock}{%
  \newenvironment{apdblock}[1]
    {\par\addvspace{\bigskipamount}\noindent
     \begin{minipage}{\textwidth}\def\@captype{#1}\centering}
    {\end{minipage}\par\addvspace{\bigskipamount}}%
}{}
\makeatother

\onecolumn
\section{Supplementary Methods}\label{apd:A}

\begin{apdblock}{table}
\caption{Example of serializing tabular data into structured text for
input.}\label{tab:apdA_serialization}
\small
\begin{tabular}{@{}p{0.42\textwidth}p{0.52\textwidth}@{}}
\toprule
\bfseries Tabular Data & \bfseries Serialized Text \\
\midrule
\begin{tabular}[t]{@{}ccccc@{}}
\toprule
Age & Pregnancy & \ldots & Blood pressure & BMI \\
\midrule
39 & 1 & \ldots & 72 & 20 \\
\bottomrule
\end{tabular}
&
\parbox[t]{\linewidth}{\raggedright
Patient: \{Age\}-year-old female, \{Pregnancy\} prior pregnancy.\\[2pt]
\hspace*{1em}- Plasma glucose (2-hour OGTT): \{Glucose\} mg/dL\\
\hspace*{1em}- Blood pressure: \{Blood Pressure\} mmHg\\
\hspace*{1em}- Triceps skin fold thickness: \{Skin\} mm\\
\hspace*{1em}- Serum insulin: \{Insulin\} mu U/mL (2-hour serum)\\
\hspace*{1em}- BMI: \{BMI\} kg/m$^{2}$\\
\hspace*{1em}- Diabetes pedigree function: \{DPF\}} \\
\bottomrule
\end{tabular}
\end{apdblock}

\begin{apdblock}{table}
\caption{Example full prompt, including system prompt, patient vignette,
and the closing instruction.}\label{tab:apdA_full_prompt}
\small
\begin{tabular}{@{}p{0.94\textwidth}@{}}
\toprule
\ttfamily\raggedright
Cutting Knowledge Date: December 2023 Today Date: 26 Jul 2024\\[4pt]
You are a clinical decision support system. For each patient, classify
their risk of developing diabetes. Respond with exactly one word: HIGH or
LOW.\\[4pt]
Patient: 50-year-old female, 6 prior pregnancies.\\
- Plasma glucose (2-hour OGTT): 148 mg/dL\\
- Blood pressure: 72 mm Hg (diastolic)\\
- Triceps skin fold thickness: 35 mm\\
- Serum insulin: not measured\\
- BMI: 33.6 kg/m\textasciicircum 2\\
- Diabetes pedigree function: 0.627\\[4pt]
The cost of a false negative (missing a patient who will develop diabetes)
is equal to the cost of a false positive (flagging a patient who will not
develop diabetes). Classify this patient as HIGH or LOW risk for developing
diabetes. Respond with exactly one word.\tabularnewline
\bottomrule
\end{tabular}
\end{apdblock}

\onecolumn
\section{Implementation details for behavioral and representational extraction}\label{apd:extraction}

\subsection{Behavioral outcome construction}\label{behavioral-outcome-construction}

For each patient and prompt configuration, we performed a forward pass and extracted the next-token probabilities at the final prompt position. Because the intended HIGH and LOW responses can correspond to different tokens depending on capitalization and preceding whitespace, we predefined families of single-token variants. The HIGH family included \texttt{HIGH}, \texttt{\ HIGH}, \texttt{High}, \texttt{\ High}, \texttt{high}, and \texttt{\ high}, and the LOW family included the corresponding variants of \texttt{LOW}. The mean total probability mass was $ \geq{0.999}$ for every model, and the minimum across all the forward passes under all the conditions per model was $\geq{0.98}$. 

Let \(\mathcal{V}_{H}\) and \(\mathcal{V}_{L}\) denote the HIGH and LOW token families. We calculated the total probability assigned to each family as

\[
q_H=\sum_{v\in\mathcal{V}_{H}}p(v\mid x),
\qquad
q_L=\sum_{v\in\mathcal{V}_{L}}p(v\mid x),
\]

where \(x\) represents the complete prompt. The model's binary decision was classified as HIGH when \(q_H>q_L\) and as LOW otherwise.

Our primary continuous behavioral measure was the probability assigned to HIGH after normalizing over the two response families:

\[
P(\mathrm{HIGH}\mid\mathrm{HIGH}\cup\mathrm{LOW})
=
\frac{q_H}{q_H+q_L}.
\]

This measure isolates the model's relative preference between the two permitted responses while reducing the influence of probability assigned to unrelated vocabulary tokens. As a secondary measure, we calculated the HIGH-versus-LOW family log-odds:

\[
\Delta_{\mathrm{logit}}
=
\log(q_H)-\log(q_L).
\]

The two continuous measures preserve the same ordering of model preferences but use different scales. The normalized probability ranges from zero to one, whereas the log-odds measure is centered at zero, with positive values indicating a preference for HIGH and negative values indicating a preference for LOW.

\subsection{Operating-characteristic calculations}\label{operating-characteristic-calculations}

We compared the binary HIGH/LOW decision with the observed diabetes outcome for each patient. A true positive was defined as a patient with diabetes for whom the model selected HIGH, whereas a false positive was defined as a patient without diabetes for whom the model selected HIGH. For each model and trade-off specification, we calculated

\[
\mathrm{TPR}
=
\frac{\mathrm{TP}}
{\mathrm{TP}+\mathrm{FN}},
\qquad
\mathrm{FPR}
=
\frac{\mathrm{FP}}
{\mathrm{FP}+\mathrm{TN}}.
\]

Each trade-off specification therefore produced one \((\mathrm{FPR},\mathrm{TPR})\) operating point. We displayed these points in ROC space to show how changing the stated relative costs of false-negative and false-positive decisions shifted the model's operating behavior. Because the classification threshold was not varied within each trade-off specification, these points represent operating points rather than complete ROC curves.

\subsection{Residual-stream activation extraction}\label{residual-stream-activation-extraction}

We registered forward hooks on every decoder block and performed one forward pass for each patient and prompt configuration. These hooks recorded the model's activations without altering its input, internal computation, or output. At decoder layer \(\ell\), we extracted the block-output residual-stream activation at the final prompt-token position:

\[
\mathbf{h}^{(\ell)}_{i,c}\in\mathbb{R}^{d},
\]

where \(i\) indexes patients, \(c\) indexes prompt configurations, and \(d\) is the model's hidden-state dimension. Activations were recorded after the decoder block's residual update and before any standardization conducted for the probing analyses.

This procedure produced one \(d\)-dimensional vector for every layer, patient, and prompt configuration. We retained activations from all decoder layers and analyzed each layer separately, producing a layerwise probe-performance trajectory rather than selecting a single layer. The behavioral outcome and residual-stream activations were obtained from the same forward pass, ensuring that each activation corresponded directly to the computation producing the recorded HIGH-versus-LOW probabilities. When required for a probing analysis, standardization parameters were estimated using only the corresponding training data and then applied to the held-out data.

\subsection{Token-level attribution}\label{token-level-attribution}

We used gradient-based salience maps to illustrate how individual prompt tokens contributed to the model's relative preference for HIGH versus LOW. The attribution target was the log-odds of HIGH versus LOW, calculated from the total next-token probability assigned to the corresponding token families:

\[
S(x)
=
\log
\frac{
\displaystyle\sum_{v\in\mathcal V_H}p(v\mid x)
}{
\displaystyle\sum_{v\in\mathcal V_L}p(v\mid x)
},
\]

where \(x\) denotes the complete prompt and \(\mathcal V_H\) and \(\mathcal V_L\) denote the predefined families of HIGH and LOW token variants, respectively. Positive values of \(S(x)\) indicate a preference for HIGH, whereas negative values indicate a preference for LOW.

During the forward pass, hooks recorded the token-level residual-stream activation \(\mathbf h_t^{(\ell)}\) at each token position \(t\) and decoder layer \(\ell\). We then backpropagated from \(S(x)\) to obtain its gradient with respect to each token-level activation:

\[
\mathbf g_t^{(\ell)}
=
\frac{\partial S(x)}
{\partial \mathbf h_t^{(\ell)}}.
\]

For each token and layer, we calculated a gradient--activation attribution score as

\[
A_t^{(\ell)}
=
\left\langle
\mathbf h_t^{(\ell)},
\mathbf g_t^{(\ell)}
\right\rangle.
\]

Positive attribution scores indicate that a token's representation contributes toward the model's preference for HIGH, whereas negative scores indicate a contribution toward LOW. The magnitude of the score represents the strength of the local contribution under this gradient-based approximation.

For visualization, attribution scores were displayed using a diverging color scale centered at zero. Chat-template boilerplate was omitted from the displayed maps for readability, although it remained part of the model input and computation. Appendix D Figure 1 compares salience maps at 3 different layers (0, 29, 79) for the same patient under two prompt configurations: a neutral prompt without explicit cost framing and a prompt specifying a symmetric 1:1 false-negative-to-false-positive cost trade-off. This comparison was intended as an illustrative attribution analysis rather than a population-level statistical test.

\subsection{Per-patient projection analysis}\label{per-patient-projection-analysis}

We examined whether patient-specific changes in residual-stream representations under an FN-heavy versus FP-heavy cost-framing contrast were associated with changes in the model's behavioral output. The calculations below were performed separately for each model, layer, and evaluated prompt contrast, using the same patients under both prompt configurations.

\paragraph{1. Patient-level shift.}\label{patient-level-shift}

Let \(\mathbf{a}_{P,\mathrm{FN}}^{(\ell)}\) and \(\mathbf{a}_{P,\mathrm{FP}}^{(\ell)} \in \mathbb{R}^{d}\) denote the residual-stream representations at layer \(\ell\) for patient \(P\) under the FN-heavy and FP-heavy prompts, respectively. Representations were extracted at the last prompt token, as described in the activation-extraction methods. We defined the patient-level shift as

\[
\boldsymbol{\Delta}_{P}^{(\ell)}
= \mathbf{a}_{P,\mathrm{FN}}^{(\ell)}
- \mathbf{a}_{P,\mathrm{FP}}^{(\ell)}.
\]

This paired difference captures the change in the patient's representation when the cost framing switches from FP-heavy to FN-heavy while the patient information remains fixed.

\paragraph{2. Population direction.}\label{population-direction}

For each patient, we estimated a reference direction from the mean shift of the other \(n-1\) patients and normalized it to unit length:

\[
\begin{gathered}
\overline{\boldsymbol{\Delta}}_{-P}^{(\ell)}
= \frac{1}{n-1}\sum_{P'\ne P}\boldsymbol{\Delta}_{P'}^{(\ell)},
\\[0.5em]
\mathbf{d}_{-P}^{(\ell)}
= \frac{\overline{\boldsymbol{\Delta}}_{-P}^{(\ell)}}
{\left\|\overline{\boldsymbol{\Delta}}_{-P}^{(\ell)}\right\|_2}.
\end{gathered}
\]

Here, \(n=768\), so each reference direction was estimated from 767 patients. Excluding patient \(P\) prevents that patient's shift from contributing to its own reference direction. The normalized direction is defined when the mean shift has nonzero norm.

\paragraph{3. Projection score.}\label{projection-score}

We then projected the patient's shift onto this direction:

\[
s_{P}^{(\ell)}
= \left(\boldsymbol{\Delta}_{P}^{(\ell)}\right)^\top
\mathbf{d}_{-P}^{(\ell)}.
\]

The score measures the signed component of the patient's shift along the population direction. Positive scores indicate movement in the same direction as the mean shift in other patients; negative scores indicate movement in the opposite direction. Scores near zero indicate little movement along this direction, which may reflect a small overall shift or a shift approximately orthogonal to the reference direction. Because the patient-level shift is not normalized, the score reflects both its magnitude and its directional alignment, rather than angular similarity alone.

\paragraph{4. Correlation with behavior.}\label{correlation-with-behavior}

For patient \(P\) under prompt configuration \(c\), let \(p_{P,c}(\mathrm{HIGH})\) and \(p_{P,c}(\mathrm{LOW})\) denote the summed next-token probabilities over the predefined HIGH and LOW token families. We defined the family log-odds gap as

\[
g_{P,c}
= \log p_{P,c}(\mathrm{HIGH})
- \log p_{P,c}(\mathrm{LOW}),
\]

and the behavioral change as

\[
b_P = g_{P,\mathrm{FN}}-g_{P,\mathrm{FP}}.
\]

Thus, \(b_P>0\) indicates a shift toward HIGH under the FN-heavy prompt relative to the FP-heavy prompt. This family log-odds gap is also the logit of the conditional HIGH probability after renormalizing over the HIGH and LOW families.

At each layer, we calculated Pearson's \(r\) and Spearman's \(\rho\) between \(s_P^{(\ell)}\) and \(b_P\) across all 768 patients. Pearson's correlation assesses the linear association between projection scores and behavioral changes. Spearman's correlation assesses whether patients with larger projection scores also tend to rank higher in behavioral change, without requiring a linear relationship. Positive correlations therefore indicate that stronger representational shifts along the population direction are associated with larger shifts toward HIGH.

These correlations characterize representation--behavior association; they do not establish that the estimated direction causally controls the output. Correlation alone also does not establish that patient shifts consistently align with the population direction or that behavioral changes follow the specified costs correctly; these depend on the signs and distributions of the projection scores and behavioral changes themselves.

\onecolumn
\section{Implementation details for cost-magnitude analyses}\label{apd:cost-magnitude}

\subsection{Behavioral mirror test}\label{behavioral-mirror-test}

For patient \(p\), let

\[
z_p(r)=\log\!\left(\frac{P_p(\mathrm{HIGH}\mid r)}{P_p(\mathrm{LOW}\mid r)}\right)
\]

denote the family log-odds gap under FN:FP cost ratio \(r\). For the primary mirror comparison, we defined

\[
\Delta_{\mathrm{FN},p}=z_p(20{:}1)-z_p(2{:}1)
\]

and

\[
\Delta_{\mathrm{FP},p}=z_p(1{:}20)-z_p(1{:}2).
\]

The first contrast increases the relative FN penalty, whereas the second increases the relative FP penalty. Decision-theoretically correct magnitude use therefore requires

\[
\Delta_{\mathrm{FN},p}>0
\qquad\text{and}\qquad
\Delta_{\mathrm{FP},p}<0.
\]

For each model and prompt phrasing, we reported the mean contrast and the proportion of patients exhibiting the expected sign on each side. A model--phrasing cell passed the behavioral mirror test only when a majority of patients moved in the expected direction on both sides. The signs of the two mean contrasts were also reported as descriptive summaries. This distinction is important because the sign of the mean measures the average magnitude and can be influenced by a small number of large responses, whereas the majority criterion measures how consistently the expected direction occurs across patients.

We repeated the mirror analysis for the following paired contrasts:

\begin{enumerate}
\def\labelenumi{\arabic{enumi}.}
\item
  FN 20:1 versus 2:1, paired with FP 1:20 versus 1:2;
\item
  FN 10:1 versus 2:1, paired with FP 1:10 versus 1:2; and
\item
  FN 20:1 versus 5:1, paired with FP 1:20 versus 1:5.
\end{enumerate}

Each contrast is computed within a single cost direction. Consequently, subtracting a patient-specific reference value---such as the patient's no-framing log-odds---from every condition does not change the contrast. For any patient-specific reference \(a_p\),

\[
\begin{aligned}
&\{z_p(r_{\mathrm{large}})-a_p\}\\
&\quad-\{z_p(r_{\mathrm{small}})-a_p\}\\
&=z_p(r_{\mathrm{large}})-z_p(r_{\mathrm{small}}).
\end{aligned}
\]

The mirror test is therefore invariant to whether the behavioral grid is expressed in absolute log-odds or relative to the no-framing baseline.

\subsection{Differential and common-mode decomposition}\label{differential-and-common-mode-decomposition}

A response on only one side cannot distinguish cost-sensitive magnitude use from a generic response to larger numerals. We therefore decomposed the paired FN- and FP-side contrasts into differential and common-mode components:

\[
m_p=\frac{\Delta_{\mathrm{FN},p}-\Delta_{\mathrm{FP},p}}{2},
\]

\[
c_p=\frac{\Delta_{\mathrm{FN},p}+\Delta_{\mathrm{FP},p}}{2}.
\]

The differential component \(m_p\) captures mirrored movement: it is positive when the FN contrast moves toward HIGH and the FP contrast moves toward LOW. The common-mode component \(c_p\) captures movement shared by the two sides and therefore reflects a response that is insensitive to which error is more costly. We additionally calculated

\[
D_p=\frac{|m_p|}{|c_p|}
\]

as a descriptive dominance ratio. Larger values indicate that the differential component is larger than the common-mode component. Because this ratio becomes unstable or undefined when \(|c_p|\) is close to zero, we interpreted it together with the unnormalized differential and common-mode components rather than as a standalone pass criterion.

\subsection{Behavioral coherence test}\label{behavioral-coherence-test}

The mirror test uses selected pairs of ratios and can therefore be satisfied even if the model responds irregularly at intermediate ratios. To evaluate within-side ordering, we computed a separate Kendall rank correlation for each patient and cost direction.

On the FN side, the ratios were ordered from 2:1 to 20:1. Moving along this sequence increases the FN penalty and should increase the HIGH-versus-LOW log-odds. On the FP side, the ratios were ordered from 1:20 to 1:2. Moving along this sequence decreases the FP penalty and should also increase the HIGH-versus-LOW log-odds. Thus, although the two sides represent opposing cost directions, their orderings were intentionally defined so that a cost-correct response yields positive rank concordance on both.

For each patient \(p\), we calculated

\[
\tau_{\mathrm{FN},p}
=
\tau_b\!\left(\log R_{\mathrm{FN}}, z_p(R_{\mathrm{FN}})\right)
\]

across the five FN-side ratios and

\[
\tau_{\mathrm{FP},p}
=
\tau_b\!\left(\log R_{\mathrm{FP}}, z_p(R_{\mathrm{FP}})\right)
\]

across the five FP-side ratios, where the side-specific values of \(R\) follow the orderings described above. Kendall's \(\tau_b\) was used because it evaluates rank concordance and accounts for tied values. A patient exhibited the two-side positive signature when

\[
\tau_{\mathrm{FN},p}>0
\qquad\text{and}\qquad
\tau_{\mathrm{FP},p}>0.
\]

We summarized the patient-level signs separately for the FN and FP sides using one-sided sign tests in both tails. Testing both tails allowed us to distinguish evidence of ordering in the hypothesized direction from evidence of systematic ordering in the opposite direction. We also reported the proportion of patients with the two-side positive signature.

Because each side contained only five cost ratios and tied model responses could occur, exact permutation probabilities were obtained using a tie-safe enumeration implemented for this analysis. This avoided relying on an asymptotic approximation at \(n=5\) and accommodated cases for which standard exact implementations do not accept tied inputs.

We did not pool the ten FN- and FP-side observations into a single correlation. Such a pooled analysis would be driven primarily by the overall separation between FN-heavy and FP-heavy prompts and could appear strongly ordered even when responses within one or both sides were irregular. Separate side-specific correlations directly evaluate the intended property: orderly adjustment as cost magnitude changes within a fixed cost direction.

\subsection{Representational mirror test for cost magnitude}\label{representational-mirror-test-for-cost-magnitude}

We applied the same mirror principle to residual-stream activations to determine whether cost magnitude was represented in a direction-sensitive form. The FN-heavy and FP-heavy cost statements used the same vocabulary and placed the same numerals in matched positions; they differed only in which error type the larger numeral modified. This design allowed us to distinguish a representation of numerical magnitude from a representation that integrated magnitude with the semantics of cost direction.

The analysis was performed separately for every magnitude contrast, model layer, and fold of a patient-stratified five-fold cross-validation split. Within a fold, we fitted one feature-wise standardization transform to the training activations pooled across all four configurations in the paired mirror comparison and applied that transform to the corresponding held-out activations. For example, the 20-versus-2 comparison pooled the 20:1, 2:1, 1:2, and 1:20 training configurations when fitting the scaler. Using one scaler for all four conditions preserved their relative geometry while preventing information from held-out patients from entering the preprocessing step.

Using the standardized training activations, we estimated separate difference-of-means directions for the FN and FP sides. For the primary 20-versus-2 comparison, these were

\[
d_{\mathrm{FN}}=\mu(20{:}1)-\mu(2{:}1)
\]

and

\[
d_{\mathrm{FP}}=\mu(1{:}20)-\mu(1{:}2),
\]

where each centroid was calculated using training patients only. We then unit-normalized the directions:

\[
u_{\mathrm{FN}}=\frac{d_{\mathrm{FN}}}{\lVert d_{\mathrm{FN}}\rVert_2},
\qquad
u_{\mathrm{FP}}=\frac{d_{\mathrm{FP}}}{\lVert d_{\mathrm{FP}}\rVert_2}.
\]

The representational common-mode and differential vectors were defined as

\[
c_{\mathrm{repr}}=\frac{u_{\mathrm{FN}}+u_{\mathrm{FP}}}{2}
\]

and

\[
m_{\mathrm{repr}}=\frac{u_{\mathrm{FN}}-u_{\mathrm{FP}}}{2}.
\]

Because this decomposition uses unit vectors, it compares the orientation of the FN- and FP-side changes rather than their raw lengths. If the residual stream responds primarily to the numerical magnitude, the two directions should be similarly oriented, producing a relatively large common-mode component and a small differential component. If the representation incorporates cost direction correctly, increasing the FN and FP penalties should produce oppositely oriented changes, producing a larger differential component. Equivalently, the cosine similarity

\[
\cos\theta=u_{\mathrm{FN}}^{\mathsf T}u_{\mathrm{FP}}
\]

approaches 1 for a shared direction and decreases as the two directions diverge, approaching \(-1\) when they are oppositely aligned.

\subsubsection{Split-half reliability ceiling}\label{split-half-reliability-ceiling}

Even when two high-dimensional, finite-sample centroid differences estimate the same underlying direction, their observed cosine similarity will generally be less than 1 because of sampling variability. We therefore estimated a within-side split-half reliability ceiling. Within the training set, patients were divided into two halves; the same FN-side or FP-side contrast direction was estimated independently in each half; and the cosine similarity between the two half-sample estimates was computed. This provides an empirical reference for the similarity expected between two estimates of the same direction at the observed sample size and activation dimensionality. The FN-versus-FP cosine was interpreted relative to these within-side reliability estimates rather than relative to a theoretical ceiling of 1 alone.

\subsubsection{Held-out projection analysis}\label{held-out-projection-analysis}

We evaluated whether the differential direction generalized to unseen patients. The training-fold differential vector was oriented as \(m_{\mathrm{repr}}= (u_{\mathrm{FN}}-u_{\mathrm{FP}})/2\). Each held-out activation vector was projected onto this vector, and the resulting scalar scores were used to rank the larger- and smaller-magnitude conditions separately on the FN and FP sides. Under cost-direction encoding, the larger FN penalty should project higher than the smaller FN penalty, yielding an FN-side AUROC above 0.5. Because a larger FP penalty should move in the opposite direction, the larger FP condition should project lower than the smaller FP condition, yielding an FP-side AUROC below 0.5 under the same orientation.

Under pure numeral encoding, \(u_{\mathrm{FN}}\) and \(u_{\mathrm{FP}}\) estimate the same shared magnitude direction. Their subtraction therefore removes the shared signal, leaving a differential vector composed primarily of training-fold estimation noise. Such noise should not systematically rank magnitude conditions in held-out patients, so both differential-projection AUROCs should approach 0.5. Thus, the paired FN-above-0.5 and FP-below-0.5 pattern indicates that the residual stream contains a magnitude component whose sign depends on which error is being penalized.

\subsubsection{Cross-side transfer without refitting}\label{cross-side-transfer-without-refitting}

Finally, we tested whether a direction estimated on one cost side transferred to the other. The unit-normalized FN direction \(u_{\mathrm{FN}}\) was applied directly, without refitting, to held-out FP activations; conversely, \(u_{\mathrm{FP}}\) was applied directly to held-out FN activations. The resulting AUROCs quantify whether a magnitude direction learned from one side can rank the magnitude contrast on the other side and, critically, whether that ranking retains or reverses its orientation. A direction-agnostic numeral representation predicts same-oriented transfer across sides. A cost-sensitive representation predicts an inversion: a direction oriented toward the larger FN penalty should rank the larger FP penalty in the opposite direction, and vice versa. This transfer analysis complements the differential projection by showing whether the two side-specific directions share a general numerical component, exhibit cost-correct reversal, or contain both components.

All scalers, centroids, contrast directions, decompositions, and projection axes were estimated from training-fold observations only and then applied unchanged to held-out patients.

\subsection{Scope of inference}\label{scope-of-inference}

The behavioral mirror test evaluates whether selected increases in FN and FP costs shift the model's output in opposite, cost-correct directions. The coherence test evaluates whether responses are rank ordered across the full set of ratios within each side. The representational mirror test evaluates whether residual-stream changes distinguish cost direction rather than merely numerical magnitude. These are necessary but not sufficient properties of proportional cost sensitivity: none establishes that a change in the specified cost ratio produces a quantitatively proportional change in the model's decision threshold or output log-odds.

\onecolumn
\section{Supplementary Figures and Tables}\label{apd:D}

\makeatletter
\@ifundefined{apdblock}{%
\newenvironment{apdblock}[1]
  {\par\addvspace{\bigskipamount}\noindent
   \begin{minipage}{\textwidth}\def\@captype{#1}\centering}
  {\end{minipage}\par\addvspace{\bigskipamount}}%
}{}
\makeatother

\setcounter{figure}{0}
\setcounter{table}{0}
\renewcommand{\figurename}{Appendix~\thesection~Figure}
\renewcommand{\tablename}{Appendix~\thesection~Table}
\renewcommand{\thefigure}{\arabic{figure}}
\renewcommand{\thetable}{\arabic{table}}
\renewcommand{\theHfigure}{apdD.\arabic{figure}}
\renewcommand{\theHtable}{apdD.\arabic{table}}

\begin{apdblock}{figure}
\small
  \begin{tabular}{@{}p{0.48\textwidth}p{0.48\textwidth}@{}}
  \toprule
  (a) Prompt without the cost tradeoff framing sentence for a truly high-risk diabetic patient &
  (b) Prompt with symmetric 1:1 cost of false negative to false positive for a truly high-risk diabetic patient \\
  \midrule
  Layer 0 & Layer 0 \\
  \includegraphics[width=\linewidth]{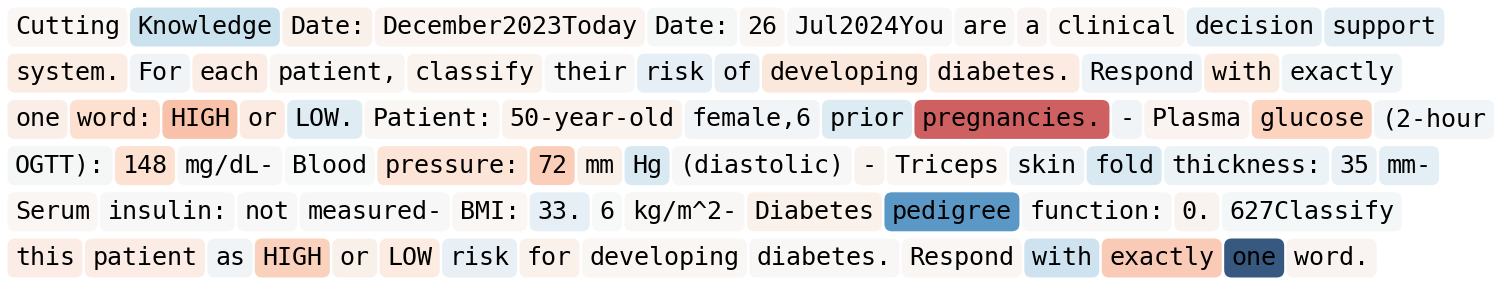} & \includegraphics[width=\linewidth]{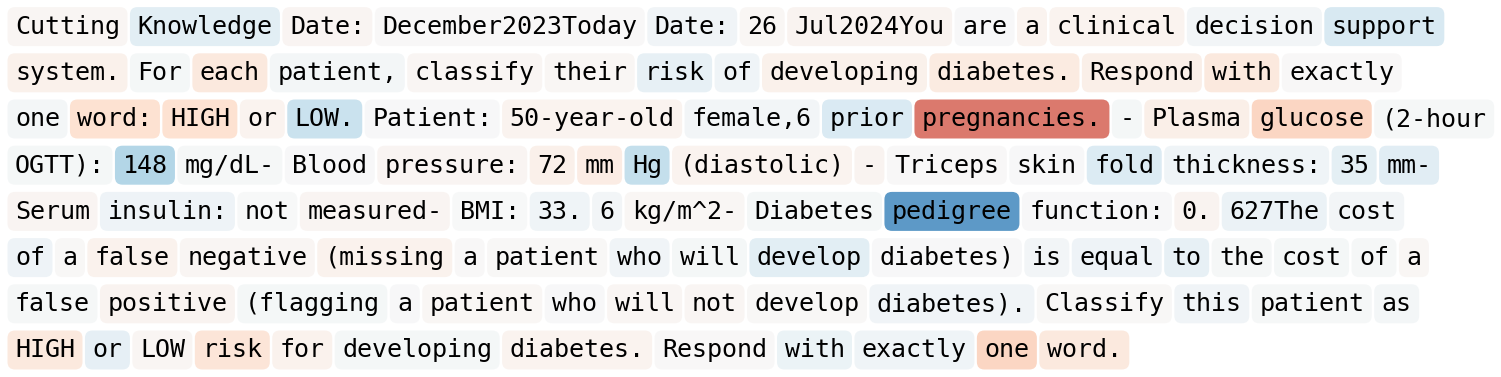} \\[4pt]
  Layer 29 & Layer 29 \\
  \includegraphics[width=\linewidth]{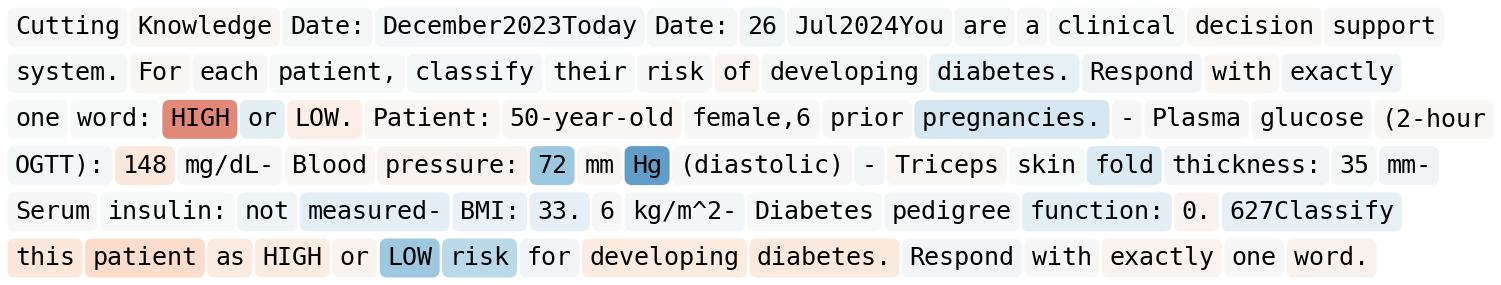} & \includegraphics[width=\linewidth]{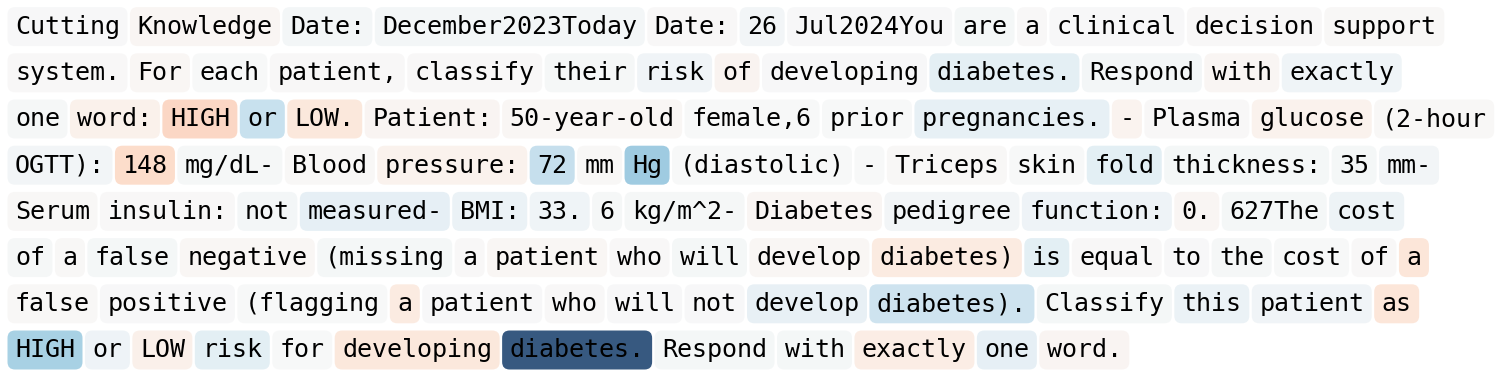} \\[4pt]
  Layer 79 & Layer 79 \\
  \includegraphics[width=\linewidth]{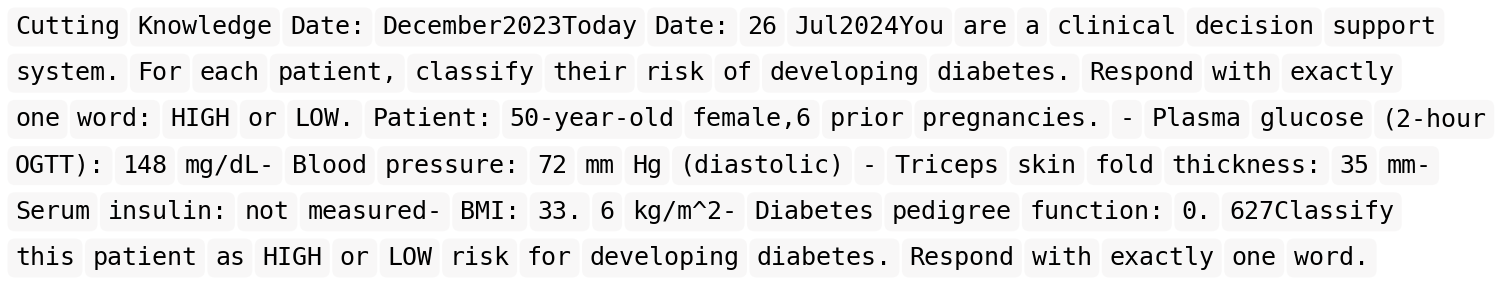} & \includegraphics[width=\linewidth]{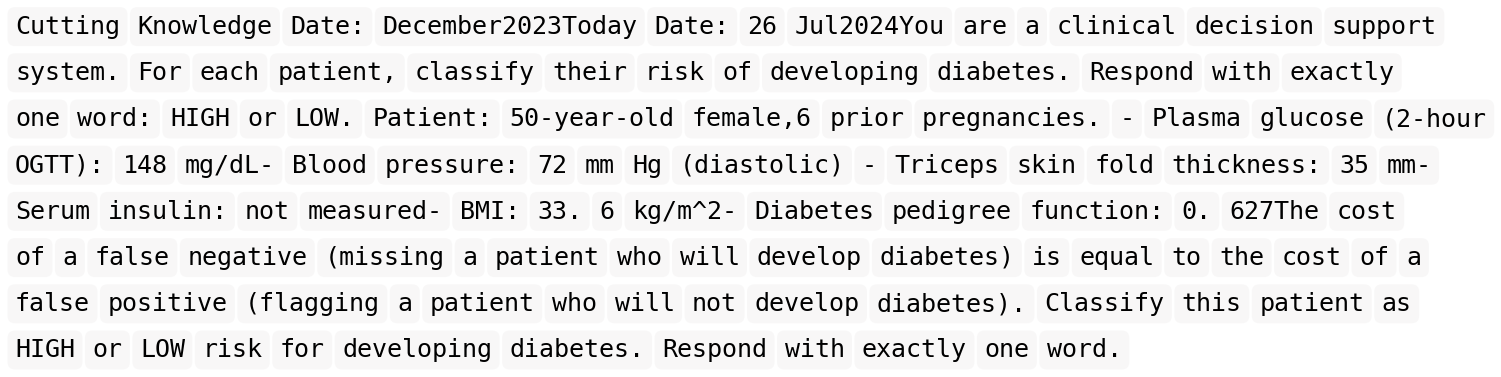} \\
  \bottomrule
  \end{tabular}
\caption{Salience maps over two set of representative prompts at 3 different layers for a high-risk patient. (a) shows examples of tradeoff-free prompt. (b) shows example prompts with symmetric 1:1 tradeoff between a false negative and a false positive for a truly high-risk diabetic patient. RED = odds in favor of HIGH, BLUE = odds in favor of LOW. Chat-template boilerplate shared across prompts was omitted from the visualization. Additional implementation details are provided in \textbf{Appendix B}.}\label{fig:apdD_salience}
\end{apdblock}

\begin{apdblock}{table}
\caption{Ground truth risk probe area under the curve (AUC) ceiling for all four models.}\label{tab:apdD_risk_auc}
\begin{tabular}{@{}lc@{}}
  \toprule
  \bfseries Model & \bfseries AUC (layer) \\
  \midrule
  Qwen-2.5-7B    & 0.830 (L18) \\
  Llama-3.1-8B  & 0.813 (L11) \\
  Qwen-2.5-32B   & 0.832 (L37) \\
  Llama-3.1-70B & 0.824 (L33) \\
  \bottomrule
  \end{tabular}
\end{apdblock}

\begin{apdblock}{figure}
\begin{tabular}{@{}cc@{}}
  Qwen-2.5-7B & Llama-3.1-8B \\
  \includegraphics[width=0.48\textwidth]{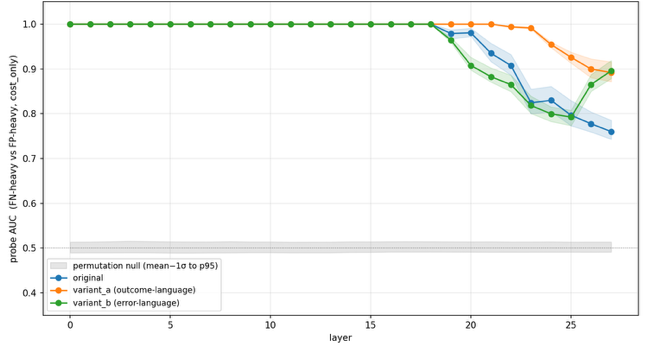} &
  \includegraphics[width=0.48\textwidth]{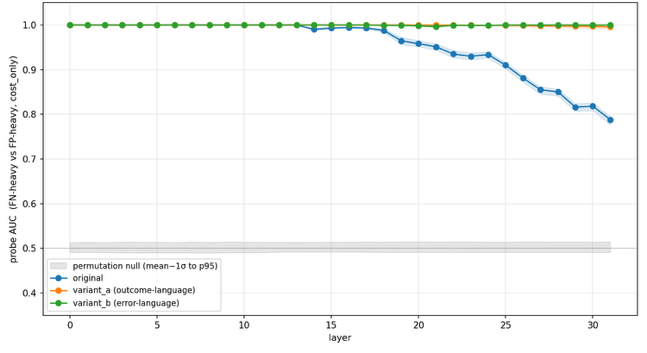} \\[4pt]
  Qwen-2.5-32B & Llama-3.1-70B \\
  \includegraphics[width=0.48\textwidth]{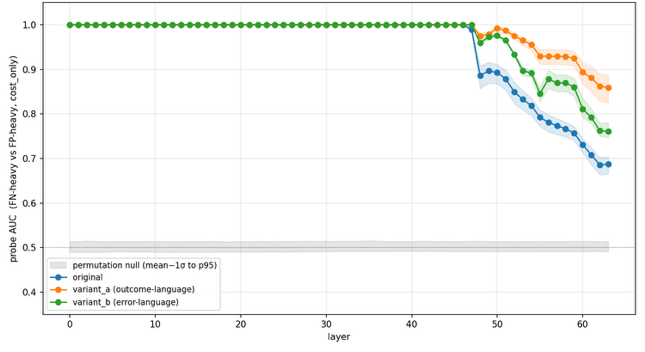} &
  \includegraphics[width=0.48\textwidth]{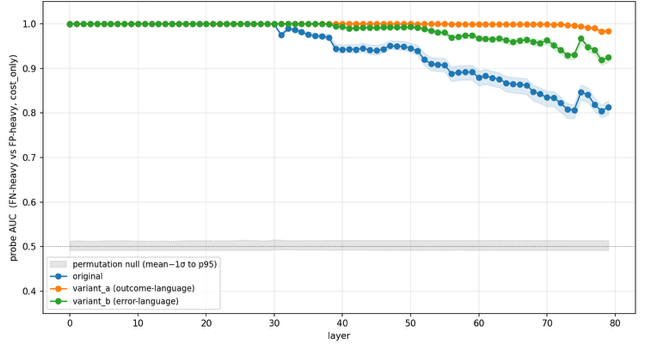} \\
  \end{tabular}
\caption{Layer-wise probe performance under paraphrase variants of the original cost tradeoff prompt for all 4 models. Variant A was paraphrased as ``Missing a patient who will develop diabetes is [X] times as costly as flagging a patient who will not develop diabetes.'' Variant B was paraphrased as ``A missed diagnosis is [X]-fold more serious than an unnecessary alert.''}\label{fig:apdD_paraphrase_probes}
\end{apdblock}

\begin{apdblock}{figure}
\begin{tabular}{@{}cc@{}}
  Qwen-2.5-7B & Qwen-2.5-32B \\
  \includegraphics[width=0.48\textwidth]{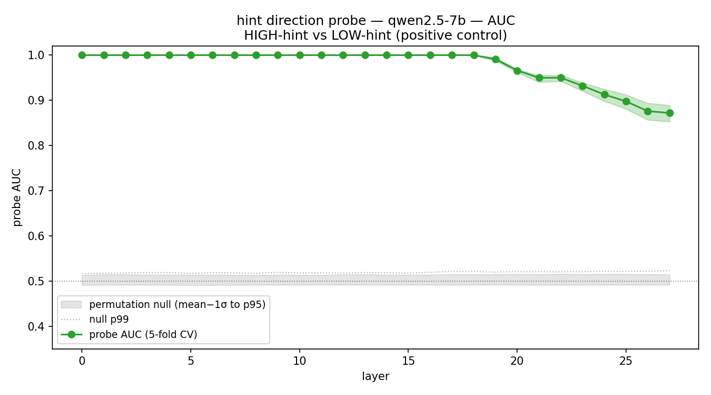} &
  \includegraphics[width=0.48\textwidth]{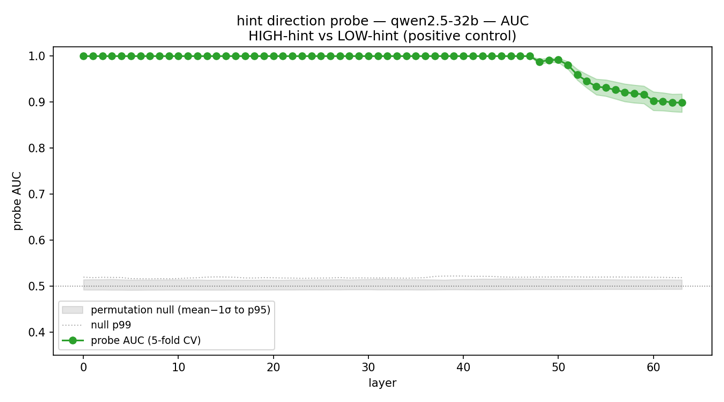} \\[4pt]
  Llama-3.1-8B & Llama-3.1-70B \\
  \includegraphics[width=0.48\textwidth]{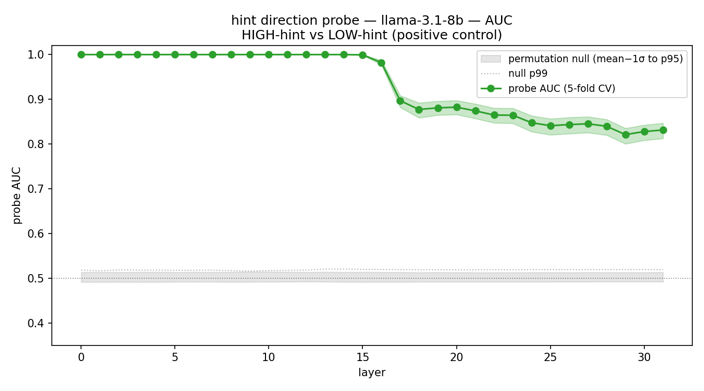} &
  \includegraphics[width=0.48\textwidth]{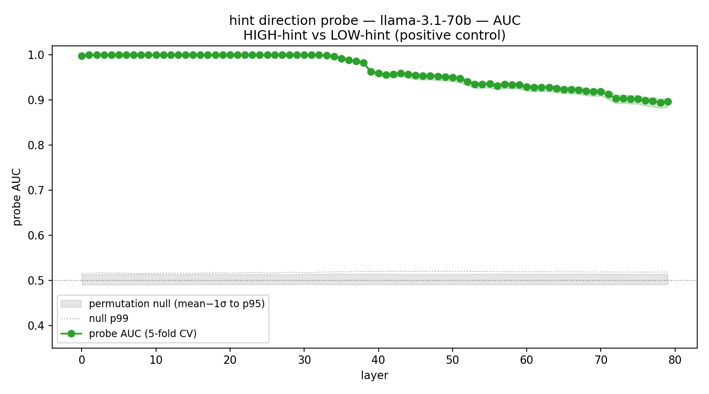} \\
  \end{tabular}
\caption{Layer-wise probe performance under \textit{cost\_plus\_hint} prompt condition for all 4 models. The \textit{cost\_plus\_hint} sentence reads ``when uncertain, err towards high/low'', depending on the cost of false negative to cost of false positive ratio.}\label{fig:apdD_hint_probes}
\end{apdblock}

\begin{apdblock}{figure}
\begin{tabular}{@{}c@{}}
  2-fold contrast between FN and FP (2:1 versus 1:2) \\
  \includegraphics[width=0.8\textwidth]{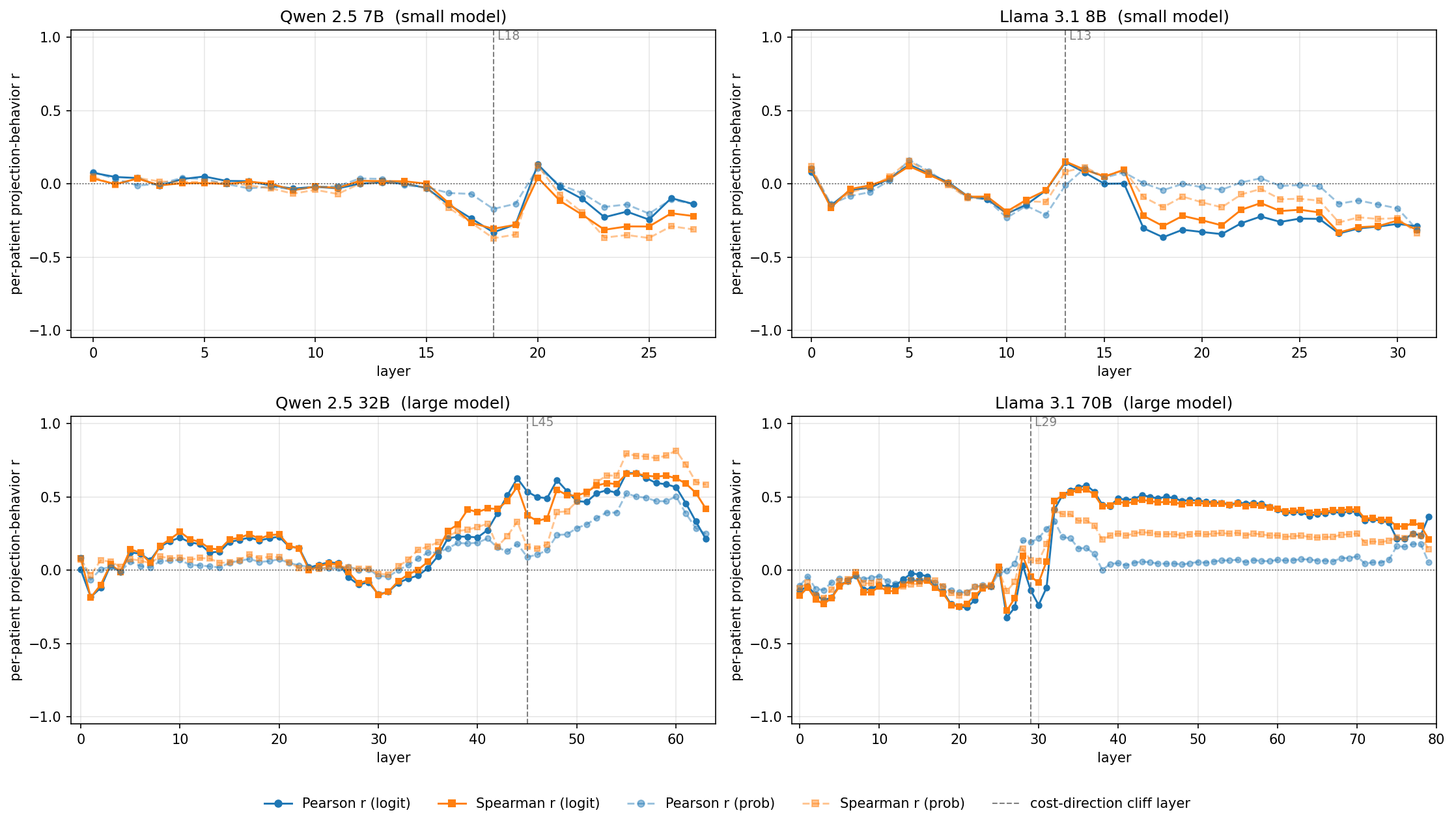} \\[6pt]
  3-fold contrast between FN and FP (3:1 versus 1:3) \\
  \includegraphics[width=0.8\textwidth]{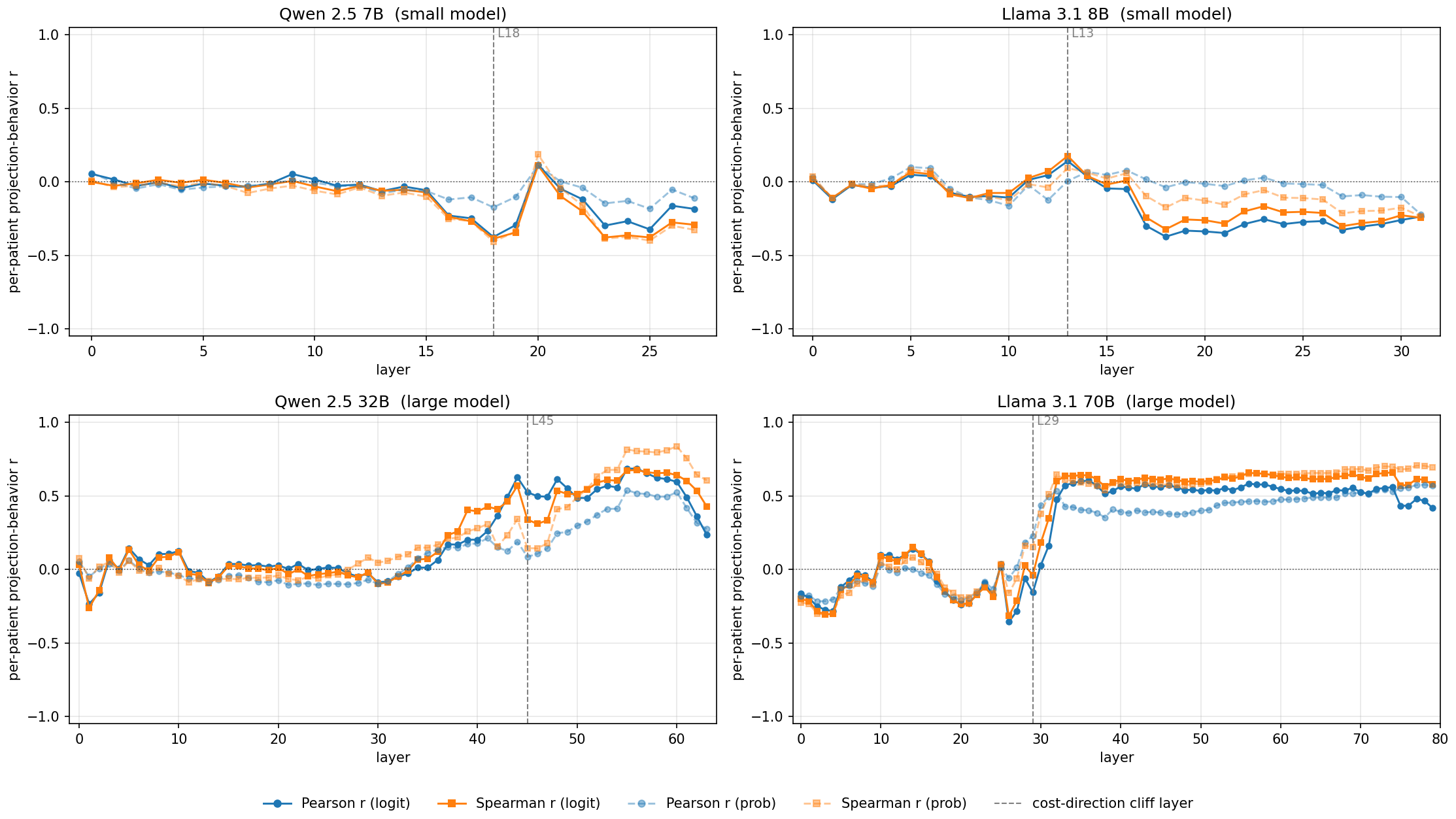}
  \end{tabular}
\end{apdblock}

\begin{apdblock}{figure}
\begin{tabular}{@{}c@{}}
5-fold contrast between FN and FP (5:1 versus 1:5) \\
\includegraphics[width=0.8\textwidth]{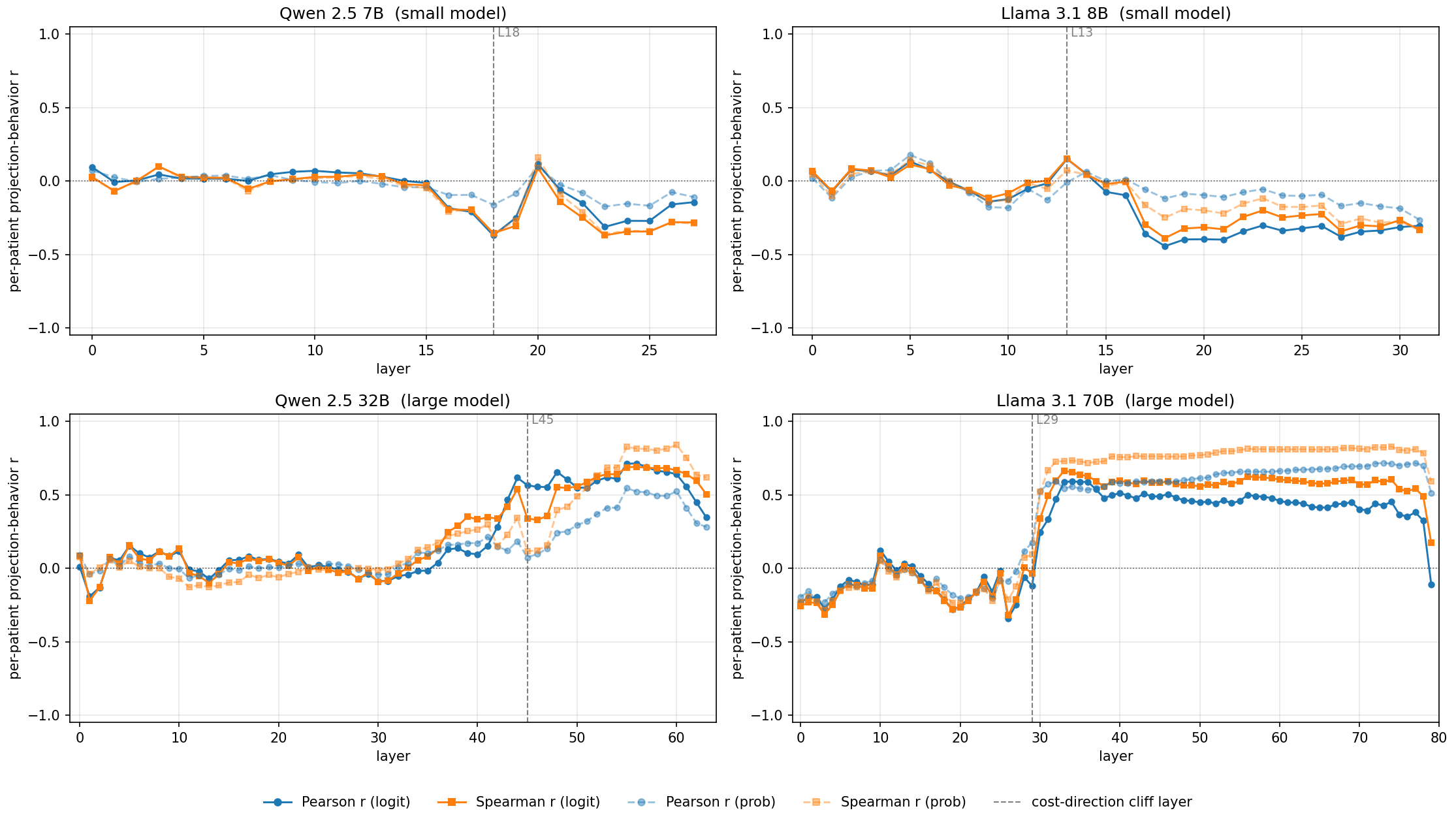} \\[6pt]
10-fold contrast between FN and FP (10:1 versus 1:10) \\
\includegraphics[width=0.8\textwidth]{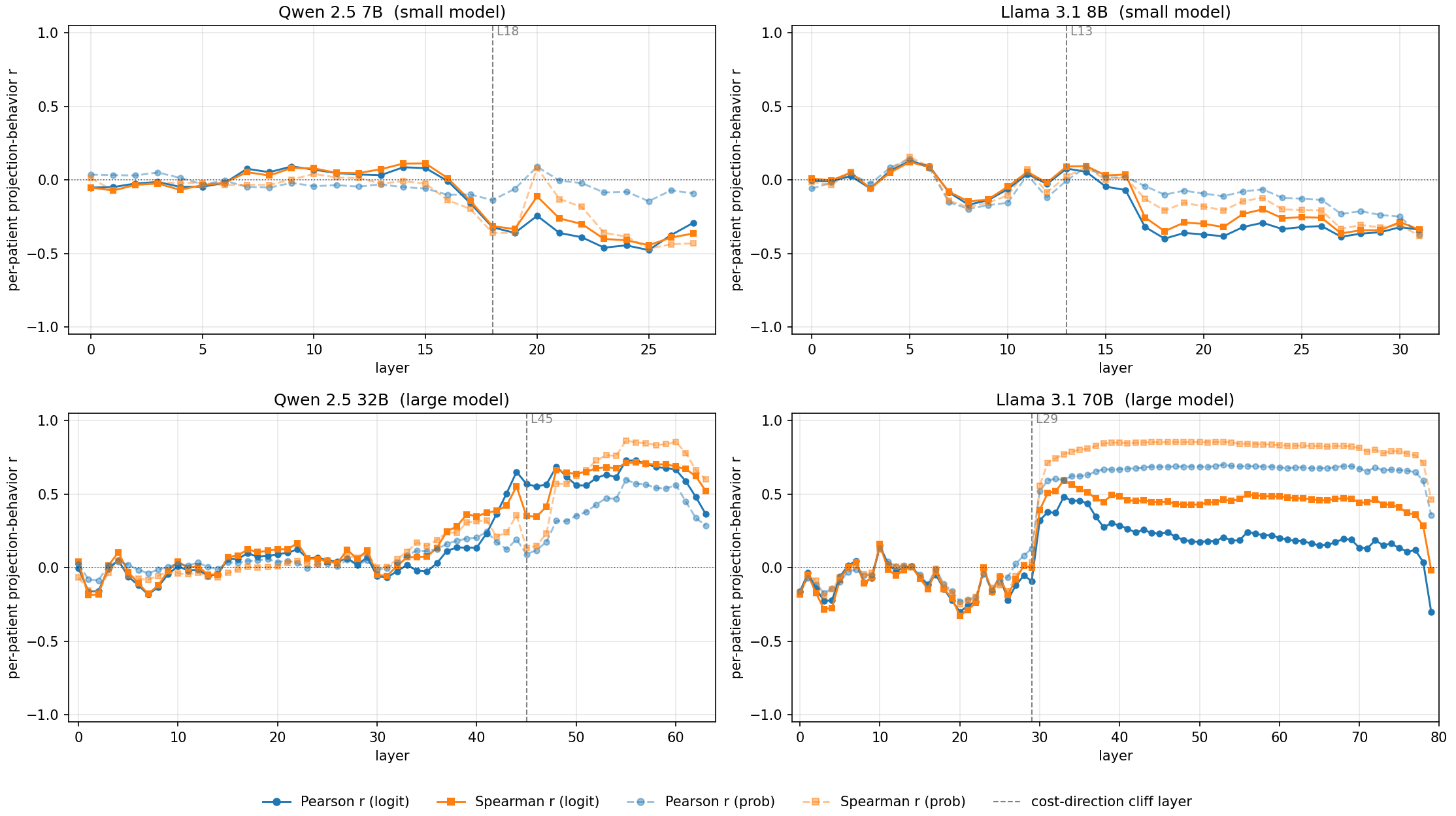}
\end{tabular}
\caption{Layer-wise per-patient activation projection versus behavioral delta per model for a 2-fold, 3-fold, 5-fold, and 10-fold contrast between FN and FP. The vertical dotted lines mark the cliff region identified in the previous linear probe trajectory. Pearson and Spearman correlations between $s(P) = \Delta(P) \cdot d_{-P}$ population-direction projection score, and $b(P) = \left( \text{logit}_{\text{HIGH}} - \text{logit}_{\text{LOW}} \right)_{\text{FN}} - \left( \text{logit}_{\text{HIGH}} - \text{logit}_{\text{LOW}} \right)_{\text{FP}}$, per-patient behavioral delta across all 768 patients. A high correlation indicates that patients whose representations shift more strongly along the cost-direction also show larger logit-gap shifts under cost-framing, that is representation predicts behavior at the individual level.}\label{fig:apdD_projection_behavior}
\end{apdblock}

\begin{apdblock}{table}
\caption{Behavioral mirror test across 4 models and 3 phrasings contrasting 20 versus 2 cost tradeoff magnitude. Cost-correct here requires positive logit delta on the FN-side and negative logit delta on the FP-side. Brackets give 95\% confidence intervals.}\label{tab:apdD_mirror_20v2}
\footnotesize\setlength{\tabcolsep}{2.4pt}%
\scriptsize\setlength{\tabcolsep}{2pt}%
  \begin{tabular}{@{}lrcrrcrcr@{}}
  \toprule
  \bfseries\begin{tabular}[b]{@{}l@{}}model /\\\quad phrasing\end{tabular} & \bfseries\begin{tabular}[b]{@{}c@{}}FN-side\\(want $+$)\end{tabular} & \bfseries\begin{tabular}[b]{@{}c@{}}Confidence\\Interval\end{tabular} & \bfseries\begin{tabular}[b]{@{}c@{}}Proportion\\(\%)\end{tabular} & \bfseries\begin{tabular}[b]{@{}c@{}}FP-side\\(want $-$)\end{tabular} & \bfseries\begin{tabular}[b]{@{}c@{}}Confidence\\Interval\end{tabular} & \bfseries\begin{tabular}[b]{@{}c@{}}Proportion\\(\%)\end{tabular} & \bfseries\begin{tabular}[b]{@{}c@{}}Differential /\\Common-mode\end{tabular} & \bfseries Dominance \\
  \midrule
  \multicolumn{9}{@{}l}{Llama-3.1-70B} \\
  \quad original & 0.713 & $[+0.685, +0.741]$ & 767/768 (99.9\%) & $-$0.244 & $[-0.265, -0.222]$ & 604/768 (78.6\%) & +0.479 / +0.235 & 2.04 \\
  \quad variant A & 0.182 & $[+0.153, +0.210]$ & 479/768 (62.4\%) & 0.036 & $[+0.016, +0.056]$ & 374/768 (48.7\%) & +0.073 / +0.109 & 0.67 \\
  \quad variant B & $-$0.142 & $[-0.155, -0.128]$ & 152/768 (19.8\%) & $-$0.217 & $[-0.227, -0.208]$ & 745/768 (97.0\%) & +0.038 / $-$0.180 & 0.21 \\
  \midrule
  \multicolumn{9}{@{}l}{Qwen-2.5-32B} \\
  \quad original & 0.459 & $[+0.436, +0.483]$ & 725/768 (94.4\%) & $-$0.672 & $[-0.706, -0.637]$ & 659/768 (85.8\%) & +0.566 / $-$0.107 & 5.31 \\
  \quad variant A & 0.151 & $[+0.133, +0.169]$ & 569/768 (74.1\%) & 0.256 & $[+0.228, +0.284]$ & 170/768 (22.1\%) & $-$0.052 / +0.203 & 0.26 \\
  \quad variant B & $-$1.450 & $[-1.488, -1.413]$ & 2/768 (0.3\%) & 0.083 & $[+0.059, +0.106]$ & 287/768 (37.4\%) & $-$0.767 / $-$0.684 & 1.12 \\
  \midrule
  \multicolumn{9}{@{}l}{Llama-3.1-8B} \\
  \quad original & 0.074 & $[+0.067, +0.081]$ & 681/768 (88.7\%) & 0.115 & $[+0.108, +0.122]$ & 49/768 (6.4\%) & $-$0.021 / +0.094 & 0.22 \\
  \quad variant A & 0.328 & $[+0.316, +0.341]$ & 719/768 (93.6\%) & 0.218 & $[+0.206, +0.229]$ & 116/768 (15.1\%) & +0.055 / +0.273 & 0.20 \\
  \quad variant B & 0.138 & $[+0.125, +0.150]$ & 598/768 (77.9\%) & 0.115 & $[+0.104, +0.126]$ & 137/768 (17.8\%) & +0.012 / +0.127 & 0.09 \\
  \midrule
  \multicolumn{9}{@{}l}{Qwen-2.5-7B} \\
  \quad original & 0.268 & $[+0.253, +0.283]$ & 698/768 (90.9\%) & 0.272 & $[+0.253, +0.290]$ & 82/768 (10.7\%) & $-$0.002 / +0.270 & 0.01 \\
  \quad variant A & 0.115 & $[+0.082, +0.146]$ & 605/768 (78.8\%) & 0.043 & $[+0.015, +0.070]$ & 313/768 (40.8\%) & +0.036 / +0.079 & 0.46 \\
  \quad variant B & $-$0.415 & $[-0.444, -0.386]$ & 58/768 (7.6\%) & $-$0.514 & $[-0.548, -0.480]$ & 741/768 (96.5\%) & +0.050 / $-$0.465 & 0.11 \\
  \bottomrule
  \end{tabular}
\end{apdblock}

\begin{apdblock}{table}
\caption{Behavioral mirror test across 4 models and 3 phrasings contrasting 10 versus 2 cost tradeoff magnitude. Cost-correct here requires positive logit delta on the FN-side and negative logit delta on the FP-side. Brackets give 95\% confidence intervals.}\label{tab:apdD_mirror_10v2}
\footnotesize\setlength{\tabcolsep}{2.4pt}%
\scriptsize\setlength{\tabcolsep}{2pt}%
  \begin{tabular}{@{}lrcrrcrcr@{}}
  \toprule
  \bfseries\begin{tabular}[b]{@{}l@{}}model /\\\quad phrasing\end{tabular} & \bfseries\begin{tabular}[b]{@{}c@{}}FN-side\\(want $+$)\end{tabular} & \bfseries\begin{tabular}[b]{@{}c@{}}Confidence\\Interval\end{tabular} & \bfseries\begin{tabular}[b]{@{}c@{}}Proportion\\(\%)\end{tabular} & \bfseries\begin{tabular}[b]{@{}c@{}}FP-side\\(want $-$)\end{tabular} & \bfseries\begin{tabular}[b]{@{}c@{}}Confidence\\Interval\end{tabular} & \bfseries\begin{tabular}[b]{@{}c@{}}Proportion\\(\%)\end{tabular} & \bfseries\begin{tabular}[b]{@{}c@{}}Differential /\\Common-mode\end{tabular} & \bfseries Dominance \\
  \midrule
  \multicolumn{9}{@{}l}{Llama-3.1-70B} \\
  \quad original & 0.611 & $[+0.591, +0.632]$ & 768/768 (100.0\%) & $-$0.206 & $[-0.220, -0.191]$ & 651/768 (84.8\%) & +0.408 / +0.203 & 2.01 \\
  \quad variant A & 0.283 & $[+0.264, +0.302]$ & 687/768 (89.5\%) & $-$0.029 & $[-0.047, -0.010]$ & 390/768 (50.8\%) & +0.156 / +0.127 & 1.22 \\
  \quad variant B & 0.087 & $[+0.071, +0.102]$ & 532/768 (69.3\%) & $-$0.548 & $[-0.561, -0.536]$ & 768/768 (100.0\%) & +0.317 / $-$0.231 & 1.38 \\
  \midrule
  \multicolumn{9}{@{}l}{Qwen-2.5-32B} \\
  \quad original & 0.634 & $[+0.605, +0.664]$ & 753/768 (98.0\%) & $-$0.802 & $[-0.833, -0.773]$ & 725/768 (94.4\%) & +0.718 / $-$0.085 & 8.49 \\
  \quad variant A & 0.224 & $[+0.202, +0.246]$ & 609/768 (79.3\%) & 0.329 & $[+0.301, +0.358]$ & 136/768 (17.7\%) & $-$0.053 / +0.277 & 0.19 \\
  \quad variant B & $-$0.850 & $[-0.880, -0.820]$ & 11/768 (1.4\%) & $-$0.454 & $[-0.478, -0.431]$ & 693/768 (90.2\%) & $-$0.198 / $-$0.652 & 0.30 \\
  \midrule
  \multicolumn{9}{@{}l}{Llama-3.1-8B} \\
  \quad original & 0.045 & $[+0.037, +0.052]$ & 579/768 (75.4\%) & 0.070 & $[+0.063, +0.077]$ & 154/768 (20.1\%) & $-$0.013 / +0.058 & 0.22 \\
  \quad variant A & 0.223 & $[+0.212, +0.234]$ & 692/768 (90.1\%) & 0.211 & $[+0.200, +0.221]$ & 105/768 (13.7\%) & +0.006 / +0.217 & 0.03 \\
  \quad variant B & 0.085 & $[+0.074, +0.095]$ & 514/768 (66.9\%) & 0.116 & $[+0.105, +0.127]$ & 177/768 (23.0\%) & $-$0.016 / +0.100 & 0.16 \\
  \midrule
  \multicolumn{9}{@{}l}{Qwen-2.5-7B} \\
  \quad original & 0.140 & $[+0.126, +0.153]$ & 623/768 (81.1\%) & 0.102 & $[+0.083, +0.121]$ & 191/768 (24.9\%) & +0.019 / +0.121 & 0.16 \\
  \quad variant A & $-$0.038 & $[-0.071, -0.006]$ & 512/768 (66.7\%) & $-$0.183 & $[-0.211, -0.157]$ & 575/768 (74.9\%) & +0.073 / $-$0.111 & 0.66 \\
  \quad variant B & $-$0.228 & $[-0.250, -0.206]$ & 141/768 (18.4\%) & $-$0.284 & $[-0.309, -0.258]$ & 673/768 (87.6\%) & +0.027 / $-$0.256 & 0.11 \\
  \bottomrule
  \end{tabular}
\end{apdblock}

\begin{apdblock}{table}
\caption{Behavioral mirror test across 4 models and 3 phrasings contrasting 20 versus 5 cost tradeoff magnitude. Cost-correct here requires positive logit delta on the FN-side and negative logit delta on the FP-side. Brackets give 95\% confidence intervals.}\label{tab:apdD_mirror_20v5}
\footnotesize\setlength{\tabcolsep}{2.4pt}%
\scriptsize\setlength{\tabcolsep}{2pt}%
  \begin{tabular}{@{}lrcrrcrcr@{}}
  \toprule
  \bfseries\begin{tabular}[b]{@{}l@{}}model /\\\quad phrasing\end{tabular} & \bfseries\begin{tabular}[b]{@{}c@{}}FN-side\\(want $+$)\end{tabular} & \bfseries\begin{tabular}[b]{@{}c@{}}Confidence\\Interval\end{tabular} & \bfseries\begin{tabular}[b]{@{}c@{}}Proportion\\(\%)\end{tabular} & \bfseries\begin{tabular}[b]{@{}c@{}}FP-side\\(want $-$)\end{tabular} & \bfseries\begin{tabular}[b]{@{}c@{}}Confidence\\Interval\end{tabular} & \bfseries\begin{tabular}[b]{@{}c@{}}Proportion\\(\%)\end{tabular} & \bfseries\begin{tabular}[b]{@{}c@{}}Differential /\\Common-mode\end{tabular} & \bfseries Dominance \\
  \midrule
  \multicolumn{9}{@{}l}{Llama-3.1-70B} \\
  \quad original & 0.370 & $[+0.356, +0.384]$ & 768/768 (100.0\%) & $-$0.100 & $[-0.113, -0.086]$ & 570/768 (74.2\%) & +0.235 / +0.135 & 1.74 \\
  \quad variant A & 0.107 & $[+0.093, +0.122]$ & 561/768 (73.0\%) & $-$0.007 & $[-0.020, +0.005]$ & 432/768 (56.3\%) & +0.057 / +0.050 & 1.15 \\
  \quad variant B & $-$0.030 & $[-0.039, -0.021]$ & 300/768 (39.1\%) & $-$0.046 & $[-0.054, -0.037]$ & 508/768 (66.1\%) & +0.008 / $-$0.038 & 0.20 \\
  \midrule
  \multicolumn{9}{@{}l}{Qwen-2.5-32B} \\
  \quad original & 0.422 & $[+0.400, +0.445]$ & 714/768 (93.0\%) & $-$0.454 & $[-0.481, -0.427]$ & 674/768 (87.8\%) & +0.438 / $-$0.016 & 27.6 \\
  \quad variant A & 0.261 & $[+0.243, +0.278]$ & 676/768 (88.0\%) & 0.287 & $[+0.262, +0.312]$ & 137/768 (17.8\%) & $-$0.013 / +0.274 & 0.05 \\
  \quad variant B & $-$0.897 & $[-0.927, -0.866]$ & 4/768 (0.5\%) & 1.800 & $[+1.739, +1.860]$ & 28/768 (3.6\%) & $-$1.349 / $-$0.452 & 2.98 \\
  \midrule
  \multicolumn{9}{@{}l}{Llama-3.1-8B} \\
  \quad original & 0.104 & $[+0.098, +0.111]$ & 695/768 (90.5\%) & 0.125 & $[+0.119, +0.132]$ & 54/768 (7.0\%) & $-$0.010 / +0.115 & 0.09 \\
  \quad variant A & 0.200 & $[+0.189, +0.212]$ & 649/768 (84.5\%) & 0.097 & $[+0.089, +0.105]$ & 219/768 (28.5\%) & $+$0.051 / +0.149 & 0.35 \\
  \quad variant B & 0.188 & $[+0.176, +0.200]$ & 618/768 (80.5\%) & 0.116 & $[+0.105, +0.126]$ & 136/768 (17.7\%) & +0.036 / +0.152 & 0.24 \\
  \midrule
  \multicolumn{9}{@{}l}{Qwen-2.5-7B} \\
  \quad original & 0.076 & $[+0.060, +0.091]$ & 570/768 (74.2\%) & 0.123 & $[+0.104, +0.141]$ & 170/768 (22.1\%) & $-$0.024 / +0.100 & 0.24 \\
  \quad variant A & 0.116 & $[+0.089, +0.143]$ & 601/768 (78.3\%) & 0.015 & $[-0.008, +0.037]$ & 424/768 (55.2\%) & +0.051 / +0.066 & 0.78 \\
  \quad variant B & $-$0.630 & $[-0.671, -0.589]$ & 25/768 (3.3\%) & $-$0.495 & $[-0.528, -0.462]$ & 738/768 (96.1\%) & $-$0.068 / $-$0.562 & 0.12 \\
  \bottomrule
  \end{tabular}
\end{apdblock}

\begin{apdblock}{table}
\caption{Per-patient coherence test across 4 models and 3 phrasings. Given the mirror tradeoff ratio experimental grid, the full-grid evaluation will be 0, and $\tau$ must be separately positive. Positive median FP $\tau$ and FN $\tau$ indicate cost-correct on both sides. Positive median FP $\tau$ and negative median FN $\tau$ indicate more patients driven towards LOW. Positive median FN $\tau$ and negative median FP $\tau$ indicate more patients driven towards HIGH. Negative median FP $\tau$ and FN $\tau$ indicate sign inversion.}\label{tab:apdD_coherence}
\small\setlength{\tabcolsep}{2.8pt}%
  \begin{tabular}{@{}llrrlr@{}}
  \toprule
  \bfseries model & \bfseries phrasing & \bfseries\begin{tabular}[b]{@{}c@{}}FN side\\median $\tau$\end{tabular} & \bfseries\begin{tabular}[b]{@{}c@{}}FP side\\median $\tau$\end{tabular} & \bfseries quadrant & \bfseries\begin{tabular}[b]{@{}c@{}}Percent with positive\\FN and FP $\tau$\end{tabular} \\
  \midrule
  Llama-3.1-70B & original & 1.00 & 0.80 & both positive & 77.7\% \\
   & variant A & 0.46 & $-$0.20 & driven to HIGH & 17.1\% \\
   & variant B & 0.00 & 0.60 & driven to LOW & 37.8\% \\
  \midrule
  Qwen-2.5-32B & original & 0.60 & 0.74 & both positive & 84.8\% \\
   & variant A & 0.32 & $-$0.40 & driven to HIGH & 9.8\% \\
   & variant B & $-$1.00 & 0.00 & inversion & 0.1\% \\
  \midrule
  Llama-3.1-8B & original & 0.46 & $-$0.60 & driven to HIGH & 1.2\% \\
   & variant A & 0.80 & $-$0.60 & driven to HIGH & 5.2\% \\
   & variant B & 0.40 & $-$0.53 & driven to HIGH & 1.8\% \\
  \midrule
  Qwen-2.5-7B & original & 0.53 & $-$0.53 & driven to HIGH & 6.0\% \\
   & variant A & 0.40 & 0.11 & both positive & 30.7\% \\
   & variant B & $-$0.40 & 0.74 & driven to LOW & 5.6\% \\
  \bottomrule
  \end{tabular}
\end{apdblock}

\begin{apdblock}{figure}
\begin{tabular}{@{}lc@{}}
  \toprule
  Contrast & Llama-3.1-70B \\
  \midrule
  20 versus 2 &
  \begin{tabular}[c]{@{}c@{}}
    \includegraphics[width=0.46\textwidth]{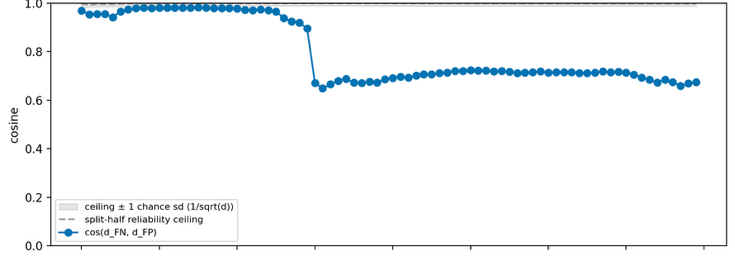} \\
    \includegraphics[width=0.46\textwidth]{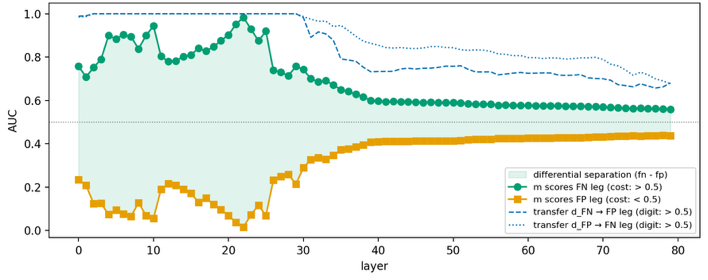}
  \end{tabular} \\
  \midrule
  10 versus 2 &
  \begin{tabular}[c]{@{}c@{}}
    \includegraphics[width=0.46\textwidth]{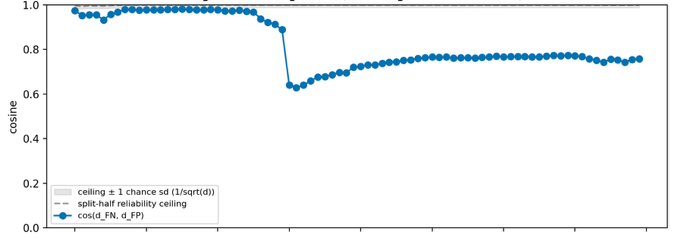} \\
    \includegraphics[width=0.46\textwidth]{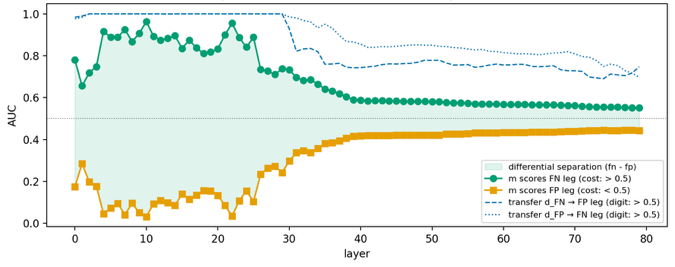}
  \end{tabular} \\
  \midrule
  20 versus 5 &
  \begin{tabular}[c]{@{}c@{}}
    \includegraphics[width=0.46\textwidth]{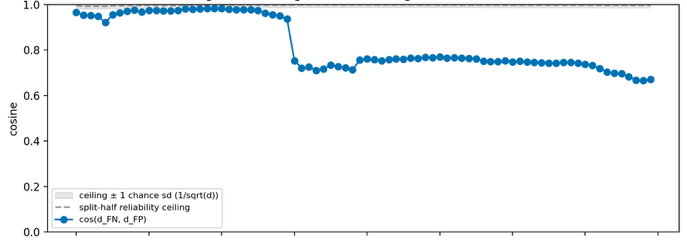} \\
    \includegraphics[width=0.46\textwidth]{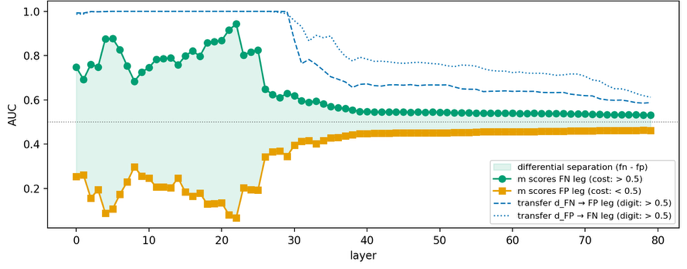}
  \end{tabular} \\
  \bottomrule
  \end{tabular}
\caption{Representational mirror test, Llama-3.1-70B for cost ratio contrasts under cost\_only prompt. Top: cosine between the FN side and the FP side difference-of-means vectors, d\_FN and d\_FP, at each layer. The dashed line is the split-half reliability ceiling and the grey band is $\pm$1 chance sd ($1/\sqrt{d}$). Bottom: held-out patient AUC for the FN (green) and FP (orange) specific differential scores (m). Dashed and dotted lines are FN and FP transfer of each difference vector applied to the numeric contrast.}\label{fig:apdD_repmirror_llama70b}
\end{apdblock}

\begin{apdblock}{figure}
\begin{tabular}{@{}lc@{}}
  \toprule
  Contrast & Llama-3.1-8B \\
  \midrule
  20 versus 2 &
  \begin{tabular}[c]{@{}c@{}}
    \includegraphics[width=0.46\textwidth]{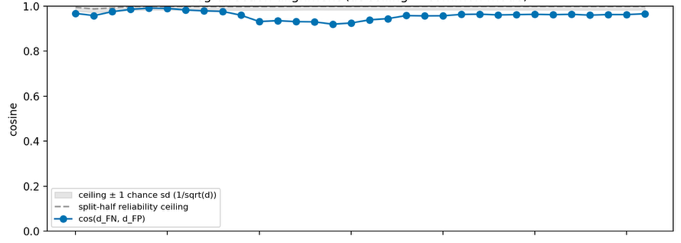} \\
    \includegraphics[width=0.46\textwidth]{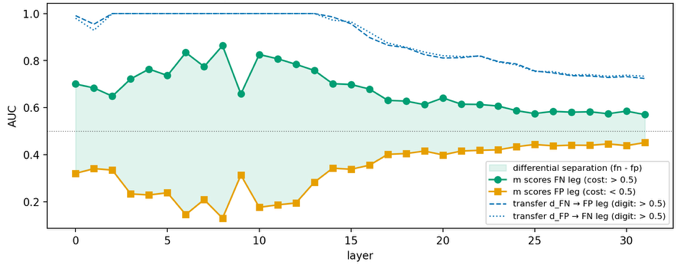}
  \end{tabular} \\
  \midrule
  10 versus 2 &
  \begin{tabular}[c]{@{}c@{}}
    \includegraphics[width=0.46\textwidth]{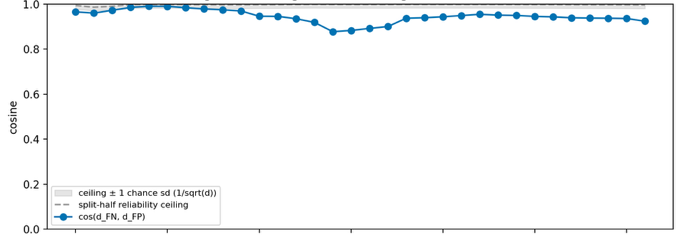} \\
    \includegraphics[width=0.46\textwidth]{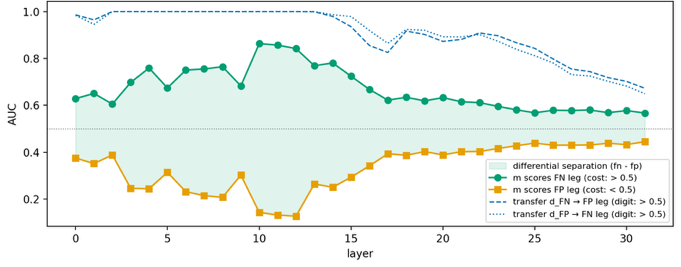}
  \end{tabular} \\
  \midrule
  20 versus 5 &
  \begin{tabular}[c]{@{}c@{}}
    \includegraphics[width=0.46\textwidth]{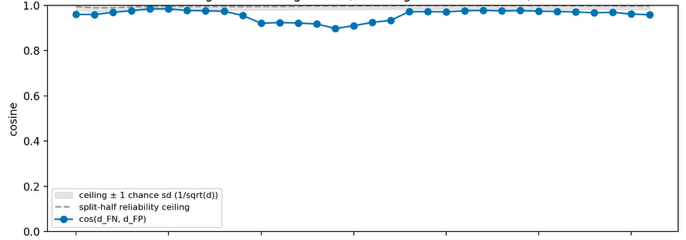} \\
    \includegraphics[width=0.46\textwidth]{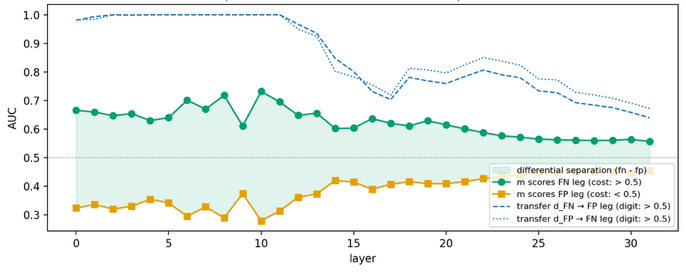}
  \end{tabular} \\
  \bottomrule
  \end{tabular}
\caption{Representational mirror test, Llama-3.1-8B for cost ratio contrasts under cost\_only prompt. Top: cosine between the FN side and the FP side difference-of-means vectors, d\_FN and d\_FP, at each layer. The dashed line is the split-half reliability ceiling and the grey band is $\pm$1 chance sd ($1/\sqrt{d}$). Bottom: held-out patient AUC for the FN (green) and FP (orange) specific differential scores (m). Dashed and dotted lines are FN and FP transfer of each difference vector applied to the numeric contrast.}\label{fig:apdD_repmirror_llama8b}
\end{apdblock}

\begin{apdblock}{figure}
\begin{tabular}{@{}lc@{}}
  \toprule
  Contrast & Qwen-2.5-32B \\
  \midrule
  20 versus 2 &
  \begin{tabular}[c]{@{}c@{}}
    \includegraphics[width=0.46\textwidth]{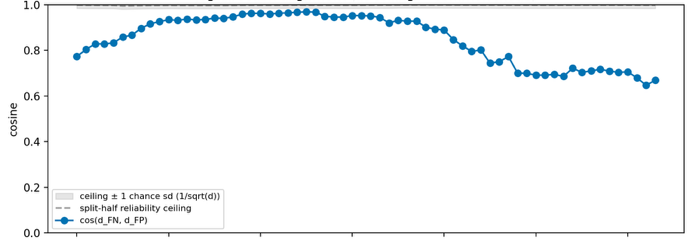} \\
    \includegraphics[width=0.46\textwidth]{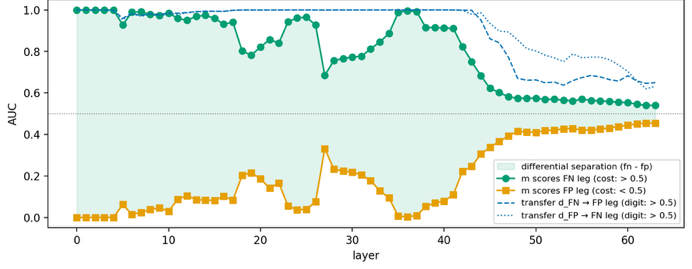}
  \end{tabular} \\
  \midrule
  10 versus 2 &
  \begin{tabular}[c]{@{}c@{}}
    \includegraphics[width=0.46\textwidth]{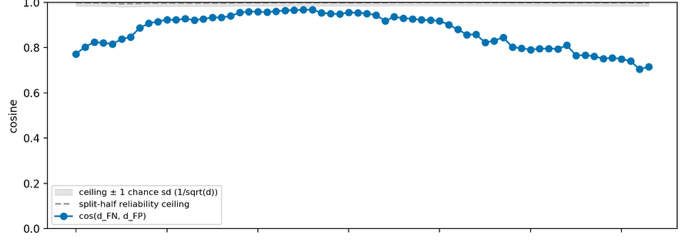} \\
    \includegraphics[width=0.46\textwidth]{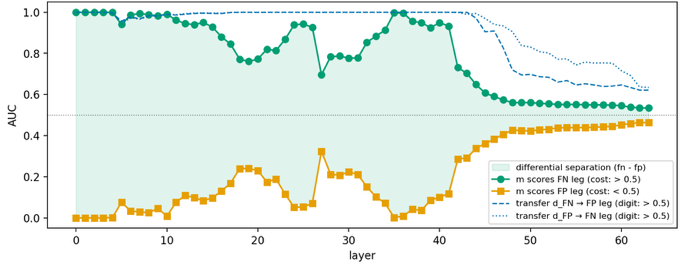}
  \end{tabular} \\
  \midrule
  20 versus 5 &
  \begin{tabular}[c]{@{}c@{}}
    \includegraphics[width=0.46\textwidth]{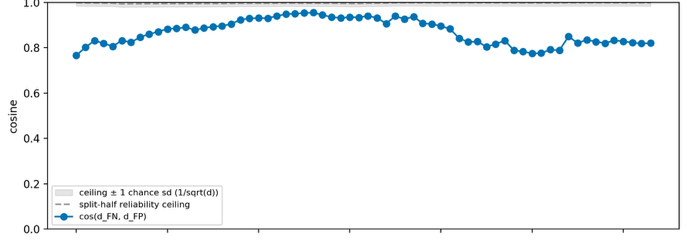} \\
    \includegraphics[width=0.46\textwidth]{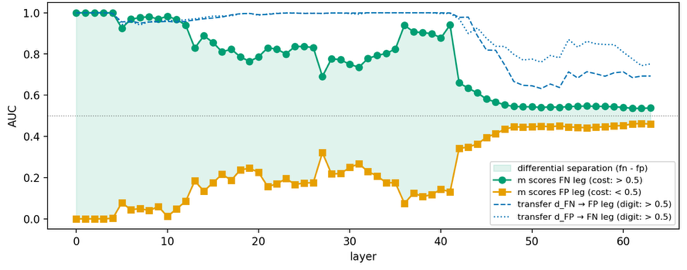}
  \end{tabular} \\
  \bottomrule
  \end{tabular}
\caption{Representational mirror test, Qwen-2.5-32B for cost ratio contrasts under cost\_only prompt. Top: cosine between the FN side and the FP side difference-of-means vectors, d\_FN and d\_FP, at each layer. The dashed line is the split-half reliability ceiling and the grey band is $\pm$1 chance sd ($1/\sqrt{d}$). Bottom: held-out patient AUC for the FN (green) and FP (orange) specific differential scores (m). Dashed and dotted lines are FN and FP transfer of each difference vector applied to the numeric contrast.}\label{fig:apdD_repmirror_qwen32b}
\end{apdblock}

\begin{apdblock}{figure}
\begin{tabular}{@{}lc@{}}
  \toprule
  Contrast & Qwen-2.5-7B \\
  \midrule
  20 versus 2 &
  \begin{tabular}[c]{@{}c@{}}
    \includegraphics[width=0.46\textwidth]{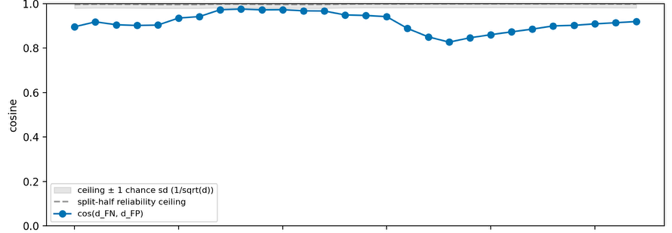} \\
    \includegraphics[width=0.46\textwidth]{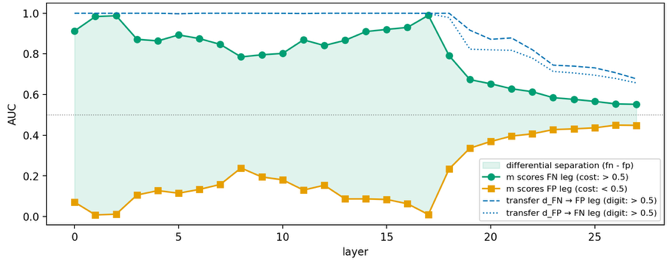}
  \end{tabular} \\
  \midrule
  10 versus 2 &
  \begin{tabular}[c]{@{}c@{}}
    \includegraphics[width=0.46\textwidth]{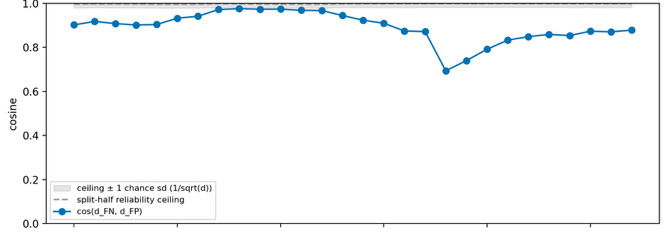} \\
    \includegraphics[width=0.46\textwidth]{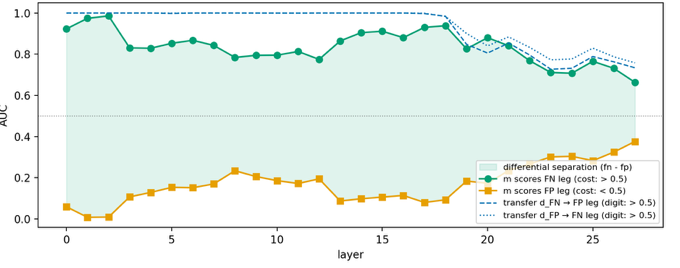}
  \end{tabular} \\
  \midrule
  20 versus 5 &
  \begin{tabular}[c]{@{}c@{}}
    \includegraphics[width=0.46\textwidth]{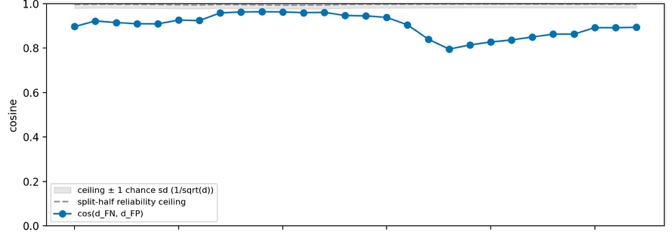} \\
    \includegraphics[width=0.46\textwidth]{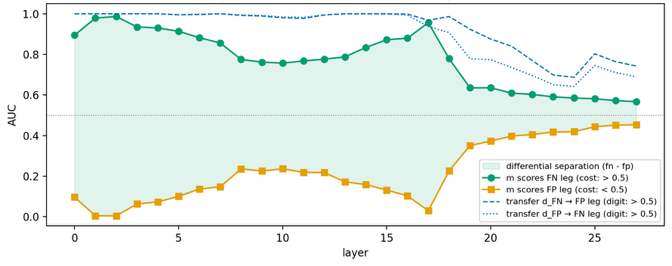}
  \end{tabular} \\
  \bottomrule
  \end{tabular}
\caption{Representational mirror test, Qwen-2.5-7B for cost ratio contrasts under cost\_only prompt. Top: cosine between the FN side and the FP side difference-of-means vectors, d\_FN and d\_FP, at each layer. The dashed line is the split-half reliability ceiling and the grey band is $\pm$1 chance sd ($1/\sqrt{d}$). Bottom: held-out patient AUC for the FN (green) and FP (orange) specific differential scores (m). Dashed and dotted lines are FN and FP transfer of each difference vector applied to the numeric contrast.}\label{fig:apdD_repmirror_qwen7b}
\end{apdblock}

\end{document}